\pdfoutput=1

\documentclass[11pt]{article}
\usepackage[table,xcdraw]{xcolor}

\usepackage{ACL2023}

\usepackage{times}
\usepackage{latexsym}
\usepackage{enumerate}
\usepackage[T1]{fontenc}
\usepackage{booktabs}

\usepackage[utf8]{inputenc}

\usepackage{microtype}

\usepackage{inconsolata}
\usepackage{caption}
\usepackage{subcaption}

\usepackage{amsmath}
\DeclareMathOperator*{\argmax}{arg\,max}

\newcommand{\p}[1]{P_{\textmd{MLM}}\left(#1\right)}
\newcommand{\pl}[1]{\mathcal{L}\left(#1\right)}

\usepackage{graphicx}
\usepackage{multirow}

\usepackage[normalem]{ulem}
\useunder{\uline}{\ul}{}

\usepackage[utf8]{inputenc}

\usepackage[LAE, T1]{fontenc}
\usepackage[arabic, english]{babel}
\usepackage[free]{ARfonts}

\title{ProMap: \\Effective Bilingual Lexicon Induction via Language Model Prompting}

\author{\normalsize Abdellah {El Mekki} $^{1}$\thanks{\hspace{0.5em}Correspondence to abdellah.elmekki@um6p.ma,\\Work done as a visiting student at UBC.} ~ Muhammad Abdul-Mageed $^{2,3}$ ~ El Moatez Billah Nagoudi $^{2}$ 
~\\ {\bf Ismail Berrada} $^{1}$ ~ {\bf Ahmed Khoumsi} $^{4}$\\
\normalsize $^{1}$College of Computing, Mohammed VI Polytechnic University\\
\normalsize $^{2}$Deep Learning \& Natural Language Processing Group,
  The University of British Columbia\\\normalsize  $^{3}$Department of Natural Language Processing \& Department of Machine Learning, MBZUAI
  \\\normalsize  $^{4}$Department of Electrical \& Computer Engineering, University of Sherbrooke}

\begin{document}
\maketitle
\renewcommand{\thefootnote}{\fnsymbol{footnote}}
\begin{abstract}
Bilingual Lexicon Induction (BLI), where words are translated between two languages, is an important NLP task. While noticeable progress on BLI in rich resource languages using static word embeddings has been achieved. The word translation performance can be further improved by incorporating information from contextualized word embeddings. In this paper, we introduce ProMap, a novel approach for BLI that leverages the power of prompting pretrained multilingual and multidialectal language models to address these challenges. To overcome the employment of subword tokens in these models, ProMap relies on an effective \textit{padded prompting} of language models with a seed dictionary that achieves good performance when used independently. We also demonstrate the effectiveness of ProMap in re-ranking results from other BLI methods such as with aligned static word embeddings. When evaluated on both rich-resource and low-resource languages, ProMap consistently achieves state-of-the-art results. Furthermore, ProMap enables strong performance in few-shot scenarios (even with less than 10 training examples), making it a valuable tool for low-resource language translation. Overall, we believe our method offers both exciting and promising direction for BLI in general and low-resource languages in particular. ProMap code and data are available at \url{https://github.com/4mekki4/promap}.

\end{abstract}

\section{Introduction}

Bilingual Lexicon Induction (BLI) is the task of automatically constructing a bilingual lexicon or a list of word translations between two different languages~\cite{mikolov2013exploiting, artetxe-etal-2018-robust, lample2018word,patra-etal-2019-bilingual,shi-etal-2021-bilingual}.  BLI has a wide range of uses, including in Natural Language Processing (NLP) tasks such as machine translation, and multilingual information retrieval, as well as in language learning and serious games. It is also vital in building systems for low-resource languages. The majority of recent BLI research focuses on using linear \cite{mikolov2013exploiting,xing-etal-2015-normalized,artetxe-etal-2016-learning,smith2017offline} and non-linear~\cite{mohiuddin-etal-2020-lnmap} mapping-based methods to align between two languages. The standard inputs to these methods are: 1) static word embeddings (WEs) of a source language $L1$ and a target language $L2$ and 2) a seed dictionary that covers a few thousand translation pairs. 

\begin{figure}[t]
	\centering
	\includegraphics[width=\columnwidth]{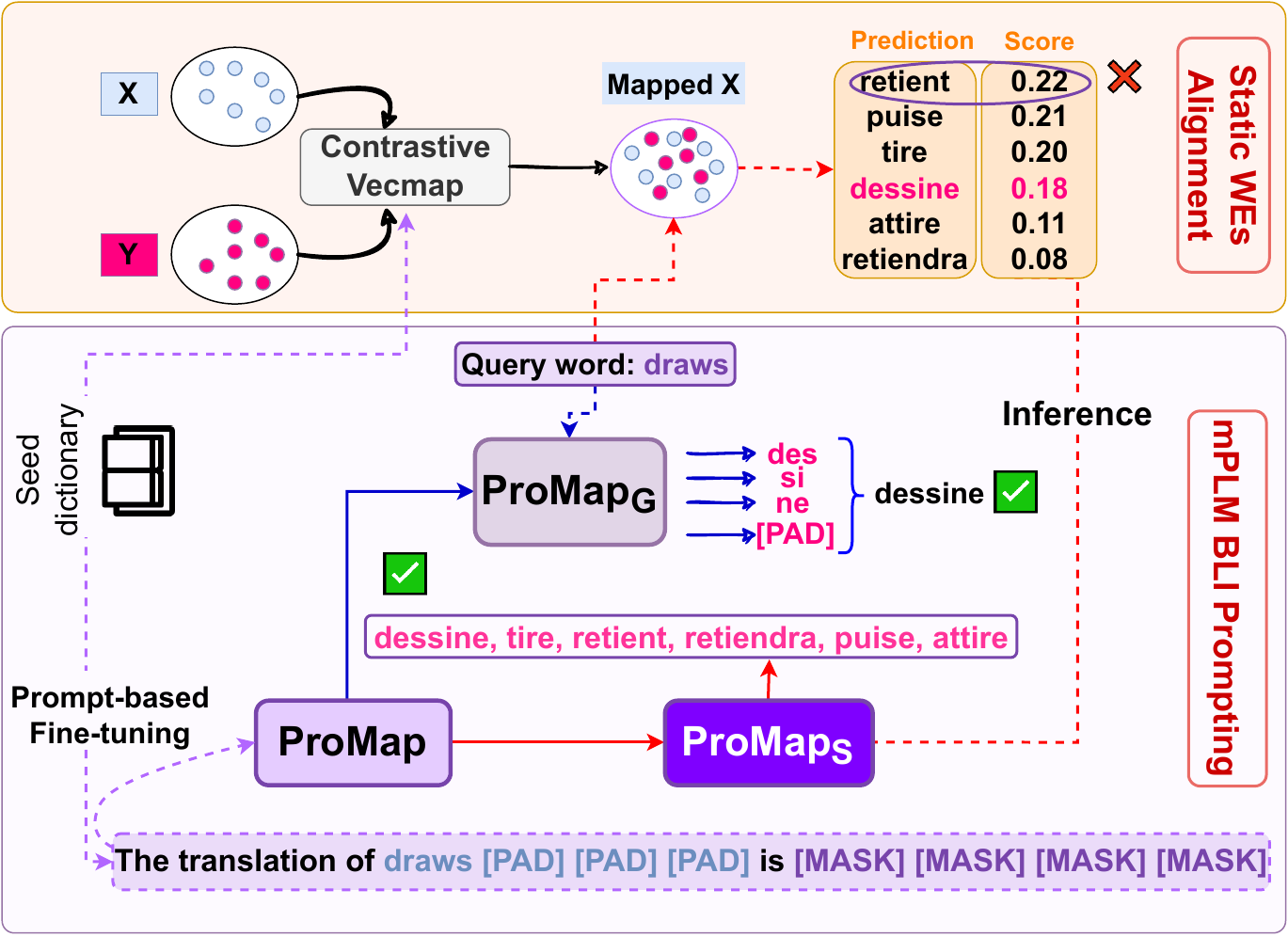}
	\vspace{-2mm}
	\caption{Overview of ProMap, depicting the workflow for translating the word "draws" from English to French. The figure illustrates the use of ProMap$_G$ for generating the translation sub-tokens and ProMap$_S$ for re-ranking Contrastive Vecamp predictions}
	\label{fig:OurScheme}
\end{figure}

Traditionally, static WEs are used to achieve state-of-the-art results in BLI. These embeddings are generated by training a language model on large amounts of monolingual texts and representing vocabulary words as points in a continuous vector space. 
For good BLI performance, the monolingual texts used to train these embeddings  should have similar distributions and come from the same domain across the source and target languages. To force this constraint, recent BLI research~\cite{artetxe-etal-2018-robust,glavas-etal-2019-properly,karan-etal-2020-classification} exploits Wikipedia dumps to train static WEs for these languages. 

Regardless, ensuring good performance in NLP tasks when resources are limited is still a known challenge. This is especially true for BLI where researchers have been criticized for studying this task using down-sampled corpora of high-resource languages \cite{artetxe-etal-2020-call} that may not be representative of real low-resource languages. For example, real low-resource languages are characterized by scripting differences, domain shifts, and a lack of sufficient bitext, resulting in less isomorphic embedding spaces and thus a decrease in BLI performance \cite{sogaard-etal-2018-limitations,nakashole-flauger-2018-characterizing,ormazabal-etal-2019-analyzing, glavas-etal-2019-properly, vulic-etal-2019-really, patra-etal-2019-bilingual, marchisio-etal-2020-unsupervised}. Arabic, a collection of diverse languages and dialect varieties, is a case in point where it is hard to find resources to build good  embeddings  for mapping-based BLI methods. This makes BLI even more challenging, especially for Arabic dialects.

Recently, BLI performance has been improved by using contextualized word embeddings~\cite{zhang-etal-2021-combining, li-etal-2022-improving}, generated from multilingual Pretrained Language Models (mPLMs), like mBERT \cite{devlin-etal-2019-bert} and XLM-R \cite{lample2019cross}. Furthermore, advancements in multilingual language modeling have led to the development of multi-dialectal models \citep{abdul-mageed-etal-2021-arbert, inoue-etal-2021-interplay}, which are designed to handle multiple dialects within a single Arabic language model. However, finetuning these PLMs on low-resource tasks can suffer from overfitting. To address this issue, prompt-based PLMs finetuning has been applied to enable few-shot learning~\cite{gao-etal-2021-making}.

In this paper, we introduce \textbf{ProMap}, a new approach of BLI that incorporates multilingual and multdialectal PLMs using \textit{padded prompt-based finetuning}. 
ProMap boosts the performance of word translation tasks and comes in two variants:
\begin{enumerate}[(i)]

    \item \textbf{ProMap$_G$} ($\textbf{G}$ for \textbf{G}eneration): This variant is particularly applicable when working with low-resource languages where it is difficult to acquire isomorphic static WEs. It directly generates the translation of a source word,  assuming the availability of a Pretrained Language Model (PLM) that can handle both the source and target languages. This variant shows promising results in \textbf{few-shot} scenarios even with less than 10 training pairs.

    \item  \textbf{ProMap$_S$} ($\textbf{S}$ for \textbf{S}election): This variant leverages an existing static WEs mapping method such as Vecamp~\cite{Artetxe_Labaka_Agirre_2018}, to select the correct translation from $K$ candidate translations proposed by this mapping.
\end{enumerate}


\noindent The main contributions of this paper can thus be summarized as follows:
\begin{enumerate}
    \item Introduction of ProMap, a novel approach for BLI  leveraging the power of pretrained multilingual and multidialectal language models. To the best of our knowledge, no prior work has tackled the BLI task using prompt-based finetuning of mPLMs.
    \item Extensive evaluation of ProMap through:  \textbf{(i)} Word translation on a standard multilingual BLI benchmark \cite{glavas-etal-2019-properly}, \textbf{(ii)} Multilingual word translation using \textit{few-shot learning}, and \textbf{(iii)}  Word translation for \textit{low-resource languages}. We evaluate word translation between Arabic dialects using four Arabic dialectal pairs following~\newcite{erdmann-etal-2018-addressing} and from ten Arabic dialects to Modern Standard Arabic (MSA) exploiting a lexicon from~\newcite{bouamor-etal-2018-madar}.
    We show that ProMap outperforms state-of-the-art methods in the majority of experiments.
\end{enumerate}

 The rest of the paper is organized as follows: Section~\ref{sec:lit} presents related work. In Section~\ref{sec:method}, we introduce our method ProMap. In Section~\ref{sec:exps}, we present our experiments and results. Section~\ref{sec:discussion} is a discussion of our results. In Section ~\ref{sec:conclusion}, we conclude the paper and draw the limitations of our work.
 
\section{Related Work}\label{sec:lit}

\paragraph{Bilingual Language Modeling.} In recent years, there has been a significant increase in research on BLI with a variety of solutions proposed \cite[e.g.][]{artetxe-etal-2016-learning, zhang-etal-2017-adversarial, zhang-etal-2017-earth, sogaard-etal-2018-limitations, patra-etal-2019-bilingual, jawanpuria-etal-2019-learning, glavas-vulic-2020-non}. On the one hand, some of these solutions, such as Procrustes-based methods, assume that the embedding spaces are roughly isomorphic. However, other researchers have argued that this assumption may not hold true \cite{patra-etal-2019-bilingual, mohiuddin-etal-2020-lnmap}, particularly for low-resource languages where it may be difficult to obtain sufficient data to construct isomorphic word embeddings \cite{feng-etal-2022-cross,graph-bli-low}. On the other hand, other studies have attempted to use BLI for translation between Arabic variants \cite{erdmann-etal-2018-addressing, El2021On}, which are considered to be very low-resource and non-standard languages \cite{SALLOUM2014372, habash-etal-2018-unified}. Nevertheless, many approaches have focused solely on high-resource languages for evaluation, which may limit advancements in the field \cite{artetxe-etal-2020-call}. Recently, there has been a trend towards combining contextualized and static word embeddings to improve alignment and boost BLI performance \cite{zhang-etal-2021-combining, li-etal-2022-improving}.

\paragraph{Prompting Pretrained Language Models.} In the recent past, more and more research has focused on the use of  prompt-based finetuning methods for language models. These studies primarily focus on identifying the most effective prompting templates \cite{schick-etal-2020-automatically, shin-etal-2020-autoprompt} and investigating the use of prompting to address few-shot learning tasks \cite{schick-schutze-2021-exploiting, schick-schutze-2021-just, gao-etal-2021-making}. Our work in this paper continues in this vein, but with particular emphasis on the task of BLI with minimal to large supervision.

\section{Method}\label{sec:method}

In this paper, we assume the availability of an mPLM trained on multiple languages or dialects. Although some languages or dialects may be low-resource in terms of non-noisy and task-specific data, we presume the availability of unlabeled data from various resources, such as social networks (e.g., Facebook, Twitter). An example of this approach is seen in MARBERT \citep{abdul-mageed-etal-2021-arbert}, which was primarily trained on 1B Arabic tweets (covering more than 20 Arab countries).

The intuitive idea of our approach for the BLI task is to finetune an mPLM by prompting it to translate a source word $w_s$ of the language $L1$ to a target word $w_t$ of the language $L2$. Although mPLMs are known for their smaller vocabulary size compared to static WEs, they tokenize words into sub-tokens. This leads to the issue of the pairs $w_s$ and $w_t$ being represented by multiple sub-tokens. To solve this problem, we introduce a padded \textbf{pro}mpting-based finetuning approach of mPLMs for word \textbf{map}ping/translation, namely \textbf{ProMap}. In summary, ProMap can be used in two variants:

\begin{enumerate}[(i)]
    \item \textbf{ProMap}$_G$: This variant is effective when there is no access to comparable static WEs while there is access to an mPLM with a reasonable vocabulary coverage. In this case, it translates $w_s$ to $w_t$ using solely the prompt-based finetuned mPLM to generate the sub-tokens that form the translation word.
    \item \textbf{ProMap}$_S$: This variant assumes the availability of both comparable static WEs and an mPLM. It uses the mPLM to re-rank the top $K$ predictions from an already existing robust alignment method between the comparable static WEs.
\end{enumerate}







\noindent Figure \ref{fig:OurScheme} summarizes an example of the use of ProMap for the translation from the word \textit{"draws"} in English to the word \textit{"dessine"} in French. The figure presents uses of both of our method's variants. In the remainder of this paper, we will refer to our method using three notations: ProMap, ProMap$_G$ and ProMap$_S$.
 We will also assume access to a training dictionary, denoted as $D_{train}$, and a testing dictionary denoted as $D_{test}$, respectively. These encompass the training and testing word pairs, respectively.

\subsection{ProMap}
\label{sec:promapmodel}

The basic idea of \textbf{ProMap} is to perform a prompt-based finetuning of the PLM, where we model the BLI task as a natural language template (more details about the prompt-based finetuning are presented in Appendix \ref{sec:method-appendix}).

 We might design our mPLM prompting template for a pair ($w_s$, $w_t$) as follows: 
\begin{equation}
\resizebox{.9\hsize}{!}{
$x_p  = \left [ CLS \right ] \text{The translation of the word}\;  w_s \; {is} \left [ MASK \right ]$}
 \label{eq:bliprom}
 \end{equation}
The masked token $[MASK]$ can be predicted using the MLM classification head of the PLM. The probability that the word $w_t$ from the PLM vocabulary $V$ will be predicted as a translation to the word $w_s$ using the template $x_p$ is:
\begin{align*}
p\left ( w_{t} \mid w_s \right ) & = p\left ( \left [ MASK \right ] = w_{t} \mid x_p \right ) \\ 
 &= softmax\left ( W_{w_{t}}\cdot h_{\left [ MASK \right ]} \right )\\ 
 &= \frac{\exp \left ( W_{w_{t}}\cdot h_{\left [ MASK \right ]} \right )}{\sum _{w_{i}\in V} \exp \left ( W_{w_{i}}\cdot h_{\left [ MASK \right ]} \right )}
\end{align*}

\noindent Where $W_{w_{*}}$ and $h_{\left [ MASK \right ]}$ refer to the hidden vectors of the target word $w_{t}$ and the $\left [ MASK \right ]$, respectively. The prompt-based finetuning utilizes the mPLM pretrained weights without adding any additional parameters, making it more efficient than standard finetuning. Thus, we can train the system by feeding all the pairs $\left ( w_{s_i}, w_{t_i} \right ) \in D_{train}$ to the mPLM model using the template $x_p$ of equation~(\ref{eq:bliprom}), and then optimizing the cross-entropy loss between the predicted $[MASK]$ value and the ground truth $w_{t_i}$.

One challenge, however, is that in $x_p$ we assume that the majority of words $w_{s_i}$ and $w_{t_i}$ are represented by a single sub-token. This assumption can be valid for PLMs that cover one (e.g. English variant of BERT) or two (e.g. GigaBERT \cite{lan-etal-2020-empirical} that covers MSA and English) languages, but for PLMs that encode a large number of languages (mPLMs) (e.g. mBERT, XLM, XLM-R), the maximum vocabulary size does not cover all words from all languages as individual sub-token each. To tackle this issue, we adapt our method to non-autoregressively predict multiple sub-tokens using padded MLM.
\\

\noindent\textbf{Padded Prompting.}
A PLM model in its original form only considers one token to be masked and infilled when using the $[MASK]$ token. To predict a span of sub-tokens of a fixed-length $n$ instead of a single token, we follow the approach of \cite{mallinson-etal-2020-felix, malmi-etal-2020-unsupervised} in using a non-autoregressive padded MLM. This approach masks a fixed-length span of $n$ tokens within a sentence and the PLM is trained to predict them while also predicting a $[PAD]$ token for masked positions that should not be infilled. 
In ProMap, we first design BLI training data for this model by converting all words $w_s$ and $w_t$ in our dictionaries $D_{train}$ and $D_{test}$ into spans of $n$ sub-tokens, padded with the token $PAD$ for words that have less than $n$ sub-tokens (the source words are also padded to unify the structure of the template over all the training examples). Then, we model our new prompt template based on the template in equation (\ref{eq:bliprom}). For example, if $n = 4$ and for the translation pair ($w_s$, $w_t$), where the sub-tokens of $w_s$ are $\left \{ w_{s_0}, w_{s_1}, w_{s_2}, w_{s_3} \right \}$, the prompt is modeled as follows:
\begin{equation}
\begin{split}
x_p =  & \left [ CLS \right ] \text{The translation of the word } \\ 
& w_{s_0} \; w_{s_1} \; w_{s_2} \; w_{s_3} \; { is }  \left [ MASK \right ] \\
& \left [MASK \right ] \; \left [ MASK \right ] \; \left [ MASK \right ]
\end{split}
\label{eq:padded-prompt}
\end{equation}

\noindent The targets to be predicted for the 4 $[MASK]$ tokens are the sub-tokens of the target word $w_{t}$ padded with $[PAD]$ to match the fixed length $n = 4$. For the training step, we follow~\newcite{malmi-etal-2020-unsupervised} in computing the pseudo-likelihood of the original sub-tokens of $w_{t}$ denoted as $W_{i:j} = w_{t_0}, w_{t_1}, w_{t_2}, w_{t_3}$ as follows:

\begin{align*}
   \pl{W_{i:j} \mid x_{p}; \Theta} = \prod_{c=i}^{j} \p{w_c \mid x_{p}; \Theta}
\end{align*}

Where $i$ and $j$ denote the range of the masked sub-tokens in $x_p$, $\p{w_c \mid x_{p}; \Theta}$ refers to the probability that the $c$-th token in $x_{p}$  takes the value $w_c$ (even a word sub-token or $[PAD]$) and $\Theta$ denotes the training data. The training of the model proceeds by finetuning the mPLM with the above formula.

\subsection{ProMap$_G$: Generation of Translation Sub-Tokens}
The first variant of ProMap, namely ProMap$_G$, predicts the translation of the source word $w_s$ based only on the mPLM model. It uses ProMap to independently generate the sub-tokens that form the predicted translation.
To get the translation of an input word $w_s$, we first pass the word through the template in equation (\ref{eq:padded-prompt}), then we decode the non-autoregressively predicted sub-tokens and concatenate them to form the prediction word.

\subsection{ProMap$_S$: Selection from $K$ Candidates}
The second variant ProMap$_S$ relies on re-ranking the predictions extracted from an existing static WEs alignment method. It uses the same finetuned ProMap model defined in section \ref{sec:promapmodel}.
\subsubsection{Static WEs Alignment}
 \label{sec:staticbli}
The objective of this step is to align the static WEs of languages $L1$ and $L2$. This is achieved by mapping both WEs into a shared embedding space through the use of dual linear mapping. This operation involves the use of two linear transformation matrices. As reported in~\newcite{Artetxe_Labaka_Agirre_2018}, a self-training process is conducted after each mapping iteration such that the training dictionary is expanded and the mapping performance is improved. In ProMap$_{S}$, we follow the method outlined in~\newcite{li-etal-2022-improving}, namely CLC1, which involves utilizing contrastive learning (CL) optimization in conjunction with self-training at each mapping iteration.

From the shared embedding space and for every source word $w_s$, we extract the top $K$ word translation candidates $P = \left [ p_1, p_2, ...,p_k \right ]$ and their corresponding similarity scores \footnote{We use cosine similarity to compute the similarity between word vectors.} $S = \left [ s_1, s_2, ...,s_k \right ]$ between every word vector $p_i \in P$ and $x_s$ (the static word vector of the $w_s$).

\subsubsection{Re-ranking $K$ Candidates}
In this step, we use the set of candidates $P$ and the finetuned ProMap model from Section \ref{sec:promapmodel} to re-rank and select the correct translation of a source word $w_s$. First, we convert the cosine similarity score vector $S$ to probability weights using $softmax$ with a standard temperature T, as follows:

\begin{equation*}
\resizebox{.9\hsize}{!}{
    $SW_i = softmax(s_i) = \frac{e^{s_{i}/T}}{\sum_{j=1}^k e^{s_{j}/T}} \ \ \ $}
\end{equation*}

\noindent Where $SW_i$ denotes the softmax score for each cosine similarity score $s_i$. Then, we compute the loss of $x_s$  as $L_{PLM} = \left [ l_{plm_1}, l_{plm_2}, ..., l_{plm_k} \right ]$, such as $l_{plm_i}$ denotes the average cross-entropy loss ($L_{ce})$ when the word $p_i$ is fed to the ProMap as translation of $x_s$. It is expressed as:

$$l_{plm_i} = \frac{1}{m}\sum_{j=0}^{m} L_{ce}(pt_j, t_j)$$

Where:

\begin{itemize}
    \item $m$ is the number of valid sub-tokens in $p_i$ (sub-tokens different from $[PAD]$).
    \item $pt_j$ and $t_j$ represent the $j$-th sub-token predicted by the MLM classifier and the $j$-th sub-token from the word $p_i$, respectively. 
\end{itemize}

\noindent Then, we compute $S_{PLM}$\footnote{In order to ensure that the scale and direction of losses are consistent with the softmax probabilities, we apply a logarithmic transformation and inverse function to the losses.}

\begin{equation*}
S_{PLM} = \left [ s_{plm_1}, s_{plm_2}, ..., s_{plm_K} \right ] 
\end{equation*}
where:
\begin{equation*}
s_{plm_i} = SW_i . \frac{1}{\log\left ( 1 + l_{plm_i} \right )}
\end{equation*}

\noindent The selected translation is $p_c \in P$ where:
\begin{equation*}
c = \argmax_{i}(s_{plm_i})
\end{equation*}
This score refers to the best token in $P$ chosen by \textbf{ProMap}$_S$.

\section{Experiments}\label{sec:exps}
We evaluate the performance of ProMap variants on two different scenarios: 1) language pairs that have access to both static WEs and mPLM, and 2) language pairs that only have access to mPLM. We use P@1 to compare our results with the baselines.

\subsection{Data}
\label{sec:data}

In the first scenario, we adopt the same BLI setup from previous studies, specifically those described in \newcite{artetxe-etal-2018-robust,glavas-etal-2019-properly,karan-etal-2020-classification}. We utilize the dataset and monolingual static WEs 
proposed by \newcite{glavas-etal-2019-properly} 
which comprise both closely related and distant languages. 
 In addition, we use the XLM-17 \cite{lample2019cross} mPLM which covers $17$ languages with a vocabulary covering $200$K tokens. However, as XLM-17 does not cover all the language pairs in the described dataset, our evaluation is performed on $15$ language pairs covered by this mPLM, including English (EN), French (FR), German (DE), Turkish (TR), Italian (IT), and Russian (RU). For the translation pairs, we use $5$K training pairs for every language pair, and $2$K pairs for testing.

In the second scenario, we evaluate the word translation between Arabic variants using ProMap$_G$ in two cases. The first case involves translation between Arabic dialects, for which we adopt the methodology of \newcite{erdmann-etal-2018-addressing,El2021On} by utilizing four Arabic dialects, namely, Maghrebin (MAG), Egyptian (EGY), Gulf (GLF), and Levantine (LEV). We utilize the dictionaries proposed by \newcite{erdmann-etal-2018-addressing} in this case. In the second case, we evaluate word translation between Arabic dialects and Modern Standard Arabic (MSA). To achieve this, we construct $10$ new dictionaries between Arabic dialects and MSA utilizing the MADAR Lexicon \cite{bouamor-etal-2018-madar} which covers $10$ Arabic variants. We split these dictionaries into Train and Test sets. The sizes of these splits are reported in table \ref{tab:data-sizes-madar} in Appendix \ref{sec:madar-insights}. We employ MARBERT \cite{abdul-mageed-etal-2021-arbert} as an mPLM since it has been shown to achieve SOTA results on many NLU tasks for Arabic dialects. Also, it has a sizeable vocabulary of $100$K tokens.

\subsection{Baseline Systems}
For the first scenario, we compare ProMap variants to $6$ strong baselines, namely,  RCSLS \cite{joulin-etal-2018-loss}, Vecmap \cite{Artetxe_Labaka_Agirre_2018}, LNMap \cite{mohiuddin-etal-2020-lnmap}, FIPP \cite{sachidananda2021filtered}, CLC1 \cite{li-etal-2022-improving} and CLC2 \cite{li-etal-2022-improving}. The first $5$ approaches only deal with static WEs, while the last one combines static WEs with contextualized WEs. For the second scenario\footnote{In the second case, we did not report any baselines due to unavailability of static WEs for the country-level Arabic variants.}, we compare our results to $4$ competitive approaches that have performed BLI work on Arabic dialects. These approaches are all based on Vecmap with several enhancements using orthographic features. A summary of each baseline system is reported in Appendix \ref{sec:appendix-baselines}.
 
\subsection{Implementation Details}
\label{sec:implementation-details}
In this work, we used Pytorch as the primary framework for building and training our models. We utilized the Huggingface library to load the pretrained models with no modifications.
Since the ProMap training requires a validation set to choose the best number of epochs, and the best hyper-parameters, we could not find a validation set for our BLI approach since the used BLI datasets lack such a set. To tackle this issue, we randomly used the language pair (EN, FR) to learn the best hyper-parameters and used them for all other language pair experiments.
We conducted experiments with different learning rates ranging from $1e$-$4$ to $5e$-$6$ and found that a learning rate of $2e$-$5$ provides the best results. The batch size was fixed to 64 for all experiments and the models were trained for 5 epochs. For the first scenario, we set the maximum length for the padded MLM to $n = 4$, the number of selected translation candidates from static WEs BLI in all experiments to $K=10$, and the temperature $T$ to $0.1$. For the second scenario, we choose $n = 1$. This indicates that the PLM will predict the translation word directly rather than multiple sub-tokens.

Table \ref{tab:trainable-parameters} in the appendix \ref{sec:implementation-details-app} presents the number of trainable parameters for each mPLM used in our paper.

\subsection{Main Results}
Table \ref{tab:scen-1-results} summarizes the main results of the multilingual experiments. For the majority of language pairs, ProMap achieves significant improvements compared to the previous SOTA methods. ProMap$_{S}$ outperforms the best static-based WEs BLI method (CLC1) by an average of $3.7$ P@1 points while outperforming the SOTA method that combines static and contextualized WEs (CLC2) by an average of $1.12$ P@1 points. It is worth mentioning that ProMap$_{S}$ improves the overall performance for both the same script (e.g. DE-FR) and different script (e.g. EN-RU) language pairs. The CLC2 baseline performs slightly better than ProMap$_{S}$ in the (DE-IT), (IT-FR), and (DE-TR) language pairs, but ProMap$_{S}$ still performs competitively in these cases. Also, ProMap$_{G}$ predicts accurate translations with the non-autoregressive generation of sub-tokens that form a whole word. It achieves $41.51$ P@1 between Italian and French words. Despite ProMap$_G$ demonstrating suboptimal performance relative to the baseline models within this context, the empirical results nonetheless indicate its effectiveness in specific applications. In particular, ProMap$_G$ exhibits proficient functionality during re-ranking processes, as demonstrated by ProMap$_S$.

\begin{table*}[hbt!]
\centering
\resizebox{2.\columnwidth}{!}{%
\begin{tabular}{lcccccc|cc}

\toprule
                             & \multicolumn{6}{c|}{\textbf{Baselines}}                                                                & \multicolumn{2}{c}{\textbf{Ours}}            \\
                             
\textbf{Pairs}               & \textbf{RCSLS} & \textbf{VecMap} & \textbf{LNMap} & \textbf{FIPP} & \textbf{CLC1} & \textbf{CLC2} & \textbf{ProMap}$_G$ & \textbf{ProMap}$_S$ \\ \midrule 
\textbf{DE $\to$ FR} & 52.74          & 50.44               & 48.46          & 50.44         & 53.78       & 55.56            & 31.47              & \textbf{56.40}           \\
\textbf{DE $\to$ IT} & 52.63          & 50.55               & 47.94          & 49.97         & 52.79       & \textbf{54.77}   & 28.19              & 54.44                   \\
\textbf{DE $\to$ RU} & 42.41          & 34.38               & 37.92          & 37.09         & 44.29       & 46.79            & 15.64              & \textbf{48.50}           \\
\textbf{DE $\to$ TR} & 30.99          & 27.18               & 29.16          & 27.65         & 34.69       & \textbf{38.86}   & 10.67              & 37.36                   \\
\textbf{EN $\to$ DE} & 57.60           & 51.00                  & 47.95          & 51.85         & 54.9        & 57.75            & 24.28              & \textbf{59.89}          \\
\textbf{EN $\to$ FR} & 66.55          & 63.10                & 62.10           & 63.25         & 65.05       & 67.20             & 46.42              & \textbf{69.38}          \\
\textbf{EN $\to$ IT} & 64.05          & 60.40                & 59.05          & 59.75         & 63.45       & 65.60             & 41.79              & \textbf{68.42}          \\
\textbf{EN $\to$ RU} & 49.40           & 39.65               & 41.10           & 42.00            & 49.15       & 50.50             & 19.57              & \textbf{54.98}          \\
\textbf{EN $\to$ TR} & 39.05          & 32.05               & 32.85          & 32.40          & 41.35       & 44.75            & 12.67              & \textbf{45.21}          \\
\textbf{IT $\to$ FR} & 66.51          & 65.89               & 64.60           & 65.32         & 66.51       & \textbf{67.86}   & 41.51              & 67.13                   \\
\textbf{RU $\to$ FR} & 47.67          & 47.51               & 43.64          & 47.15         & 50.55       & 52.70             & 30.70               & \textbf{54.06}          \\
\textbf{RU $\to$ IT} & 46.57          & 46.78               & 43.74          & 45.89         & 49.66       & 51.96            & 26.89              & \textbf{53.02}          \\
\textbf{TR $\to$ FR} & 36.10           & 36.58               & 34.08          & 34.40          & 40.63       & 43.88            & 21.23              & \textbf{43.91}          \\
\textbf{TR $\to$ IT} & 34.56          & 34.24               & 32.00             & 33.44         & 38.98       & 42.17            & 19.31              & \textbf{43.49}          \\
\textbf{TR $\to$ RU} & 28.06          & 26.20                & 26.20           & 26.36         & 32.00          & 36.16            & 11.27              & \textbf{37.17}          \\ \midrule
\textbf{Avg.}                & 47.66          & 44.40               & 43.39          & 44.46         & 49.19       & 51.77            & 25.44              & \textbf{52.89}          \\ \bottomrule
\end{tabular}}
\caption{P@1 scores on the multilingual BLI benchmark using 5K translation pairs. The highest scores among all approaches are highlighted in bold.}
\label{tab:scen-1-results}
\end{table*}

\subsection{Analyses}

\subsubsection{ProMap$_G$ vs. Static WEs Mapping}
\label{sec:promapg-effictiveness}
To demonstrate the effectiveness of ProMap$_G$, we conduct a fair comparison with other static WEs mapping approaches. To ensure the fairness of the experiments, we use the same dictionaries for training and evaluation for both ProMap$_G$ and the other approaches. Specifically, we only select word pairs that were covered by both the multilingual PLM vocabulary and the static WEs (both ProMap$_G$ and the baselines are trained on the same training pairs). The new sizes of the Train and Test dictionaries after this selection are reported in Table \ref{tab:shared-dict-results}. The results, presented in Table \ref{tab:shared-dict-results}, show that ProMap$_G$ significantly outperforms the other static WEs approaches across all $15$ language pairs with an average improvement of $10.55$ P@1 points. This is achieved for both close language pairs such as English-German, where ProMap$_G$ outperforms the best static WEs alignment method, namely CLC1, by $14.01$ P@1 points. Additionally, for distant language pairs, ProMap$_G$ shows large performance gains. This is true even for language pairs that do not share the same script, such as the Turkish-Russian pair where the performance increases from $24.66$ P@1 using the CLC1 approach to $32.88$ P@1 using ProMap$_G$, with a gain of $8.22$ P@1 points.

The impact of ProMap$_G$ on an mPLM is illustrated in in Appendix \ref{sec:visu-tsne}. The t-SNE plot \citep{JMLR:v9:vandermaaten08a} demonstrates the embeddings generated for each word in the Test sets before and after ProMap$_G$. Before finetuning, the plot shows a clear separation of the language sub-spaces by the mPLM, which explains why it is challenging to directly extract translations without finetuning. After ProMap$_G$, the shapes of the sub-spaces shift towards a shared sub-space where every token translation in the source language is projected towards its corresponding translation in the target language.

\begin{figure}[!t]
     \centering
     \begin{subfigure}[!ht]{0.49\linewidth}
         \centering
         \includegraphics[width=0.995\linewidth]{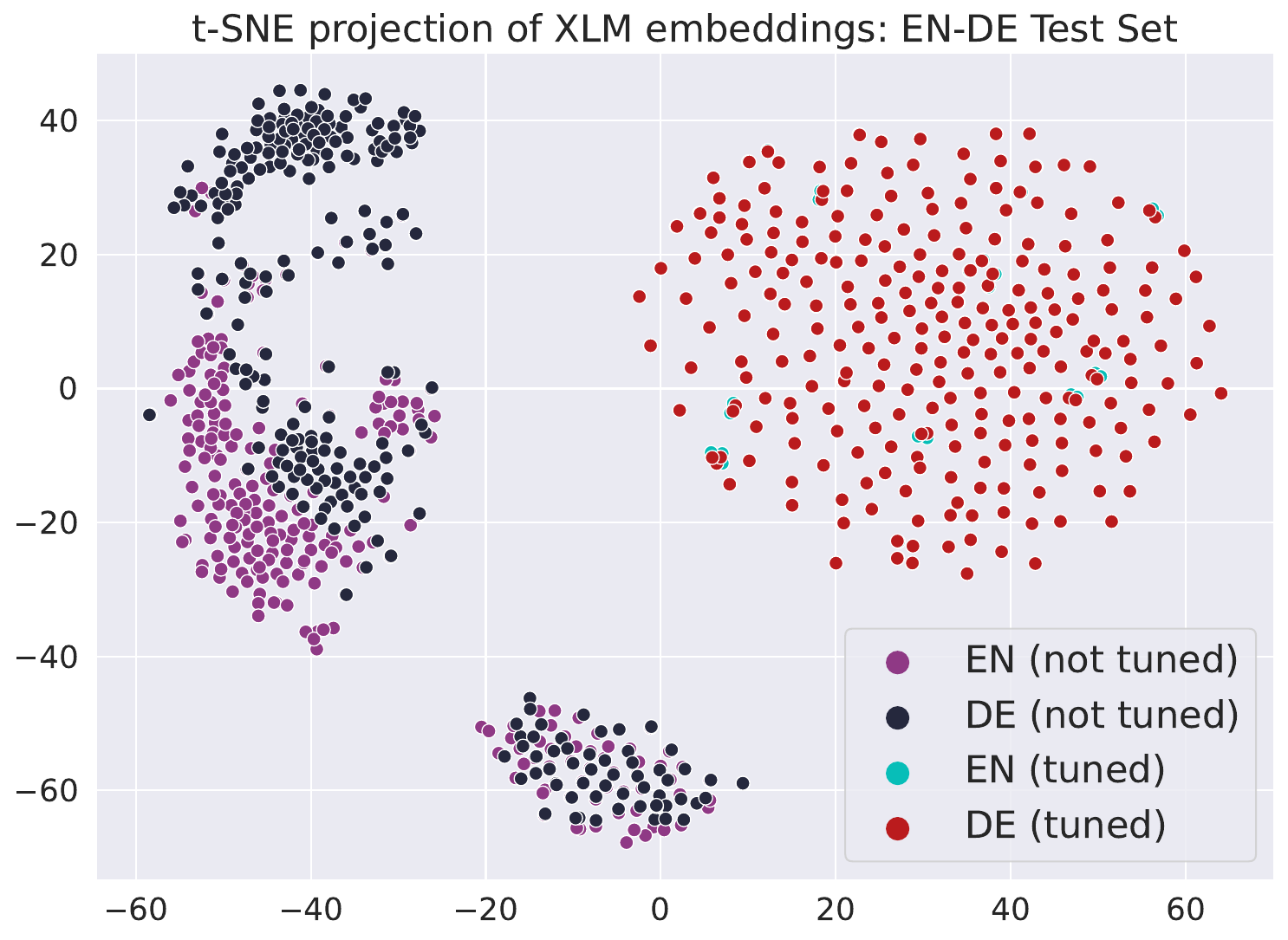}
         \label{fig:y equals x}
     \end{subfigure}
     \begin{subfigure}[!ht]{0.49\linewidth}
         \centering
         \includegraphics[width=0.995\linewidth]{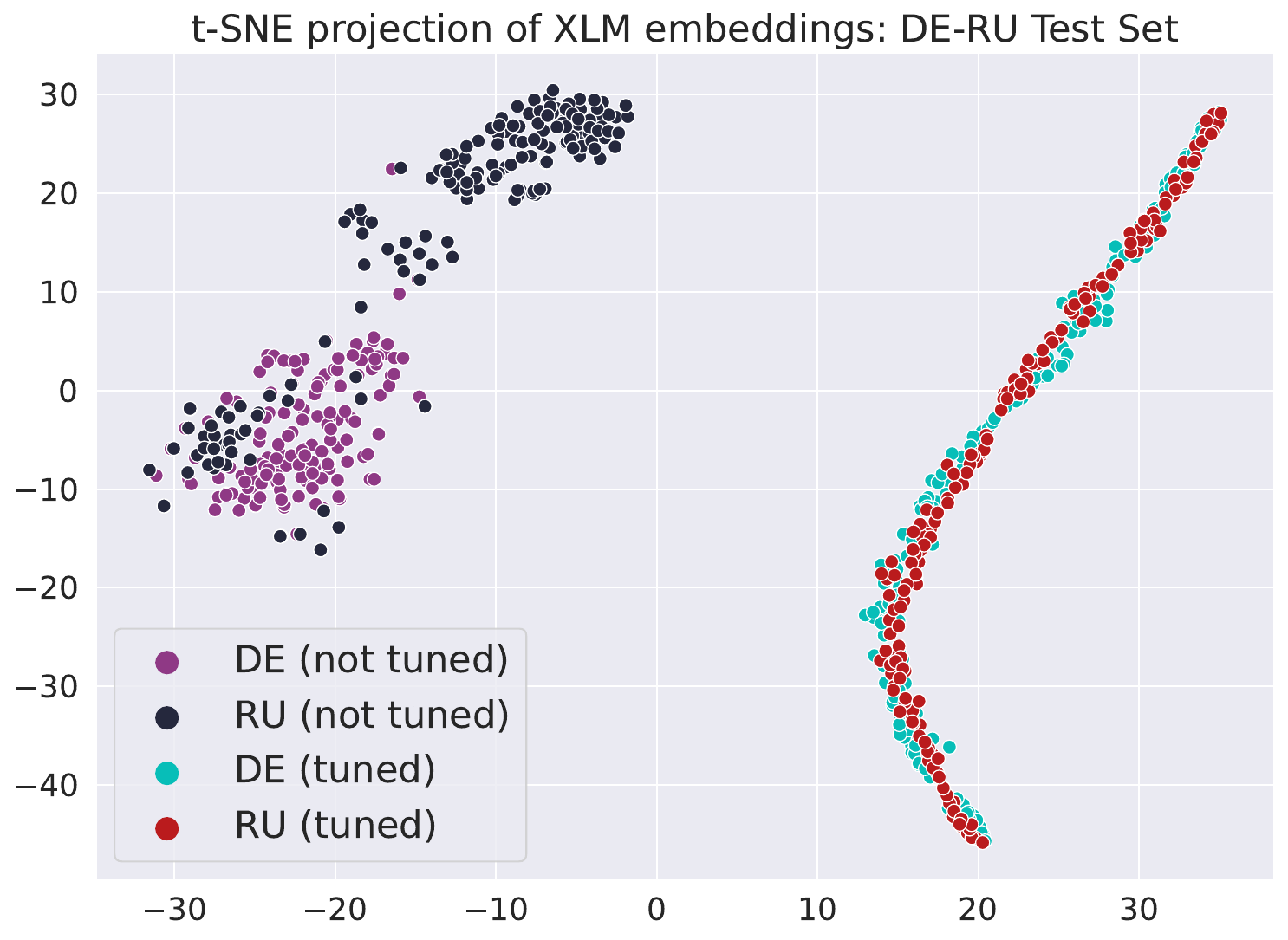}
         \label{fig:three sin x}
     \end{subfigure}
        \caption{Scatter plot of cross-lingual word embeddings for Test sets generated by the multilingual XLM model. The embeddings are represented in two dimensions using t-SNE \cite{JMLR:v9:vandermaaten08a} and depict the difference in performance before and after prompt-based finetuning.}
        \label{fig:tsne-plots-1}
\end{figure}

\begin{table}[!t]
\centering
\resizebox{\columnwidth}{!}{%
\begin{tabular}{lcccc}
\toprule
\textbf{Pairs}               & \textbf{Data Size (Train/Test)} & \textbf{VecMap} & \textbf{CLC1} & \textbf{ProMap}$_G$ \\ \midrule  
\textbf{DE $\to$ FR} & 2,189/385                        & 51.08               & 57.53       & \textbf{60.75}     \\
\textbf{DE $\to$ IT} & 2,084/368                        & 46.18               & 50.99       & \textbf{57.51}     \\
\textbf{DE $\to$ RU} & 1,256/172                        & 35.37               & 37.20       & \textbf{50.00}        \\
\textbf{DE $\to$ TR} & 1,458/248                        & 27.50               & 37.50       & \textbf{54.17}     \\
\textbf{EN $\to$ DE} & 2,180/257                        & 58.75               & 63.42       & \textbf{77.43}     \\
\textbf{EN $\to$ FR} & 2,984/361                        & 64.82               & 66.20       & \textbf{78.67}     \\
\textbf{EN $\to$ IT} & 2,797/327                        & 61.47               & 63.61       & \textbf{79.82}     \\
\textbf{EN $\to$ RU} & 1,561/131                        & 24.43               & 35.88       & \textbf{53.44}     \\
\textbf{EN $\to$ TR} & 1,871/210                        & 36.67               & 46.67       & \textbf{59.05}     \\
\textbf{IT $\to$ FR} & 2,993/586                        & 67.08               & 68.31       & \textbf{72.54}     \\
\textbf{RU $\to$ FR} & 1,651/242                        & 41.18               & 46.64       & \textbf{54.20}      \\
\textbf{RU $\to$ IT} & 1,584/231                        & 37.84               & 40.54       & \textbf{50.00}        \\
\textbf{TR $\to$ FR} & 1,894/319                        & 27.52               & 36.58       & \textbf{45.30}      \\
\textbf{TR $\to$ IT} & 1,819/311                        & 29.45               & 35.62       & \textbf{43.84}     \\
\textbf{TR $\to$ RU} & 1,086/151                        & 19.86               & 24.66       & \textbf{32.88}     \\ \midrule
\textbf{Avg.}                &  --                               & 41.95               & 47.42       & \textbf{57.97}     \\ \bottomrule
\end{tabular}}
\caption{Comparison of P@1 scores between our approach for generating word translation and static word embeddings-based alignment approaches. All models are trained on the shared pairs by the static word embeddings and the mPLM vocabularies. The highest scores among all approaches are highlighted in bold.}
\label{tab:shared-dict-results}
\end{table}

\subsubsection{Few-shot BLI with ProMap$_{G}$}
Prompt-based finetuning of PLMs has been found effective in few-shot learning tasks. 
In the following experiments, we train ProMap$_G$ with different sizes of training data between $1$ and $512$ samples. We use the same data as in Section \ref{sec:promapg-effictiveness}. For each experiment, we randomly sample $N$ pairs from $D_{train}$ and use them to train ProMap$_{G}$. We then report the result achieved on the Test dictionary. Table \ref{tab:few-shot-results} presents a subset of the results (the full results are in Table \ref{tab:few-shot-full-results} in Appendix \ref{sec:few-shot-res}). The results show that the BLI performance using ProMap$_{G}$ increases with the number of training samples. 
Additionally, ProMap$_{G}$ demonstrates promising results for word translation with minimal training examples for both closely and distantly related language pairs (such as EN-FR and RU-IT, respectively). For instance, when using only $N=1$ training example, ProMap$_{G}$ attains a P@1 score of $20.67$ for the EN-FR pair, and a score of $9.70$ for the RU-IT pair. Furthermore, with only $10$ training examples, the P@1 score for the EN-FR pair increases to $66.03$. In our experiments employing VecMap, we found that the performance was consistently 0.0 P@1 for all few-shot scenarios where N is less than 256 training examples. This suggests that while mPLMs effectively align words with their corresponding translations across languages, even when presented with a minimal number of training examples, static word embedding alignment methods such as VecMap train their embeddings independently and require substantial data to achieve comparable accuracy.

\begin{table}[]
\centering
\resizebox{\columnwidth}{!}{%
\begin{tabular}{llllll}
\toprule
\textbf{Pairs} & \textbf{DE $\to$ IT} & \textbf{EN $\to$ FR} & \textbf{EN $\to$ TR} & \textbf{IT $\to$ FR} & \textbf{RU $\to$ IT} \\ \midrule
\textbf{N=1}   & 5.34 (2.86)                  & 20.67 (15.58)                & 8.09 (7.05)                  & 13.05 (10.88)                & 9.70 (7.99)                   \\
\textbf{N=3}   & 14.99 (6.27)                 & 39.89 (8.92)                 & 16.80 (9.14)                  & 38.87 (7.81)                 & 22.40 (1.93)                  \\
\textbf{N=5}   & 29.40 (4.33)                  & 48.08 (16.50)                 & 22.48 (6.71)                 & 48.24 (8.64)                 & 25.29 (2.37)                 \\
\textbf{N=10}  & 25.27 (13.35)                & 66.03 (6.50)                  & 28.55 (11.40)                 & 52.25 (4.36)                 & 30.93 (0.94)                 \\
\textbf{N=16}  & 18.74 (6.69)                 & 65.40 (5.96)                  & 34.60 (7.35)                  & 39.57 (11.49)                & 33.10 (1.85)                  \\
\textbf{N=32}  & 34.01 (1.37)                 & 59.45 (9.21)                 & 37.68 (5.81)                 & 46.11 (6.37)                 & 36.29 (0.61)                 \\
\textbf{N=64}  & 38.34 (4.35)                 & 59.80 (10.21)                 & 41.40 (11.40)                  & 55.57 (6.67)                 & 38.40 (2.42)                  \\
\textbf{N=128} & 45.95 (2.13)                 & 70.44 (3.48)                 & 42.85 (4.34)                 & 59.06 (9.14)                 & 42.81 (2.25)                 \\
\textbf{N=256} & 50.82 (2.44)                 & 74.37 (1.98)                 & 48.94 (2.62)                 & 68.74 (2.25)                 & 45.37 (2.11)                 \\
\textbf{N=512} & 51.33 (6.86)                 & 74.21 (4.48)                 & 57.17 (7.06)                 & 66.11 (8.81)                 & 47.20 (2.21)                  \\ \bottomrule
\end{tabular}}
\caption{Comparison of P@1 Scores of ProMap$_G$ Using N-Shot training example pairs. Every value in the table presents the average (and standard deviation) of $25$ runs (corresponding to $5$ random samples x $5$ random seeds). The full results table, Table \ref{tab:few-shot-full-results}, is in Appendix \ref{sec:few-shot-res}.}
\label{tab:few-shot-results}
\end{table}

\subsection{Evaluation on Arabic Variants}
We test the effectiveness of ProMap$_G$ on Arabic variants which are considered low-resource languages.  With the limited availability of lexicons and mPLMs that cover these variants, it is hard to afford static WEs for every country-level Arabic variant. Table \ref{tab:scen-2-results1} presents the results achieved for the word translation between Arabic variants. The results show that ProMap$_{G}$ largely outperforms the baseline models by $6.90$ P@1 points on average. It is worth highlighting that we only rely on the translations directly generated using ProMap$_G$ without involving any static WEs. Also, our model outperforms the results achieved by \cite{riley-gildea-2018-orthographic, El2021On} which takes the orthographic similarities between Arabic variants into consideration when predicting the translation word. 

Furthermore, for the case of word translation between Arabic dialects and MSA (both directions), Table \ref{tab:subset-madar-results} displays a subset of the results (full results are in Table \ref{tab:madar-results}, Appendix \ref{sec:madar-full-results}). The Table presents the P@1, P@5, P@10, and P@50 scores achieved for the different pairs. On average, the performance of word translation from Arabic dialects to MSA is $58.99$ and $77.69$ for P@1 and P@5, respectively. This indicates a potential for increased transfer learning between dialects and MSA. However, the performance of word translation from MSA to dialects is lower: it has an average P@1 and P@5 scores of $40.32$ and $60.27$, respectively. This discrepancy can be attributed to the wide diversity in the dialects as the model branches out to $N$ dialects while attempting to map an MSA word to a dialectal word from the wide selection. That is, while this seems to be a one-to-one mapping between a word from MSA and a dialectal word, the model seems to be trying to learn a dialect path (from many) while selecting the target word.  

\begin{table}[hbt!]
\centering
\resizebox{\columnwidth}{!}{%
\begin{tabular}{lcccc|c}
\toprule
                                & \multicolumn{4}{c|}{\textbf{Baselines}}                                                                  & \textbf{Ours} \\
                                & \textbf{M1} & \textbf{M2} & \textbf{M3} & \textbf{M4} & \textbf{ProMap}$_G$   \\ \midrule
\textbf{EGY $\to$ GULF} & 48.30                    & 52.34                   & 53.56                     & 55.27                    & \textbf{64.40}        \\
\textbf{MAG $\to$ GULF} & 40.00                      & 44.92                   & 45.27                     & 47.87                    & \textbf{55.36}       \\
\textbf{LEV $\to$ GULF} & 41.70                    & 46.85                   & 46.03                     & 48.49                    & \textbf{60.14}       \\
\textbf{LEV $\to$ EGY}  & 37.70                    & 42.48                   & 42.52                     & 45.67                    & \textbf{57.33}       \\
\textbf{MAG $\to$ EGY}  & 36.60                    & 41.13                   & 41.96                     & 44.48                    & \textbf{55.11}       \\
\textbf{MAG $\to$ LEV}  & 54.00                      & 62.70                    & 57.01                     & \textbf{64.53}           & 55.33                \\ \midrule
\textbf{Avg.}                & 43.05                   & 48.40                   & 47.73                     & 51.05                    & \textbf{57.95}       \\ \bottomrule
\end{tabular}
}
\caption{P@1 scores on the BLI benchmark between Arabic Regions dialects following the datasets proposed by \citet{erdmann-etal-2018-addressing}. ProMap$_G$ is compared to 4 others methods namely, \textbf{M1} (\citet{erdmann-etal-2018-addressing}), \textbf{M2} (\citet{Artetxe_Labaka_Agirre_2018}), \textbf{M3} (\citet{riley-gildea-2018-orthographic}), and \textbf{M4} (\citet{El2021On}). Bold scores denote the highest scores among all approaches.}
\label{tab:scen-2-results1}
\end{table}

\begin{table}[]
\centering
\resizebox{\columnwidth}{!}{%
\begin{tabular}{lcccccccccc}
\toprule
              & \multicolumn{2}{c}{\textbf{MAR-MSA}} & \multicolumn{2}{c}{\textbf{LEV-MSA}} & \multicolumn{2}{c}{\textbf{EGY-MSA}} & \multicolumn{2}{c}{\textbf{YEM-MSA}} & \multicolumn{2}{c}{\textbf{IRQ-MSA}} \\
              & \textbf{$\to$} & \textbf{$\gets$}          &\textbf{$\to$} & \textbf{$\gets$}          & \textbf{$\to$} & \textbf{$\gets$}          & \textbf{$\to$} & \textbf{$\gets$}          & \textbf{$\to$} & \textbf{$\gets$}          \\ \midrule
\textbf{P@1}  & 52.98                     & 39.55                  & 60.37                     & 40.91                  & 61.95                     & 41.63                  & 56.72                     & 38.75                  & 67.33                     & 49.08                  \\
\textbf{P@5}  & 64.90                      & 54.80                   & 80.65                     & 64.46                  & 80.49                     & 67.81                  & 81.72                     & 59.41                  & 85.64                     & 70.18                  \\
\textbf{P@10} & 70.86                     & 59.89                  & 83.41                     & 73.14                  & 84.39                     & 74.68                  & 86.19                     & 68.63                  & 88.61                     & 75.69                  \\
\textbf{P@50} & 83.44                     & 72.32                  & 94.93                     & 87.60                   & 92.68                     & 85.41                  & 91.79                     & 82.29                  & 93.07                     & 84.86                  \\ \bottomrule
\end{tabular}}
\caption{Subset of results of ProMap$_G$ for word translation between Arabic dialects and MSA using MARBERT on the MADAR Lexicon. In the table, one country from every Arab region was selected. Table \ref{tab:madar-results} in Appendix \ref{sec:madar-full-results} presents the full results.}
\label{tab:subset-madar-results}
\end{table}

\section{Discussion} \label{sec:discussion}

Our experimental results indicate that ProMap outperforms previous BLI approaches and can generate high-quality translation pairs using both rich- and low-resource languages in both the generation and selection settings. By utilizing only the pairs covered by both the static WEs vocabulary and the mPLM vocabulary, ProMap$_G$ is able to generate largely superior results without the need to make use of, or depend on, any other approach. 
In our analysis, we find that variations in the prompting template do not have a significant effect on BLI performance (Appendix \ref{sec:effect-prompt-template}), although slightly better results are achieved when using a template written in the source language. Also, we find that injecting the source and target language information in the template does not affect the performance. Additionally, experiments show that ProMap$_G$ can learn word translation with only one training example. In addition, promising results are possible across some pairs with just ten examples.


Furthermore, regarding the diminished performance of ProMap$_G$, it is more pronounced in scenarios with multiple sub-tokens (when $n>1$). We identify two primary reasons for this. First, the multiple sub-tokens scenario can be likened to a multi-label classification task where the model is tasked with assigning multiple tags to different segments of the output. For an accurate translation in our context, all these decisions must be precise. Second, the complexity of the multiple sub-tokens scenario is exacerbated by the morphological richness of certain languages (e.g., Arabic), leading to significant variation in sub-token choices, For instance, if ProMap$_G$ employs CAMELBERT to generate the word "\AR{عالسلامة}", it must predict three sub-tokens: "\AR{عال}", "\AR{سلام}", and "\AR{ة}". Similarly, for the word "\AR{امبارح}", the model must to predict "\AR{ام}", "\AR{بار}", and "\AR{ح}". A mistake in predicting even a single sub-token can compromise the entire target translation. To mitigate this challenge, we have considered expanding the vocabulary of mPLMs by incorporating non-covered words and initializing their embedding weights by averaging the weights of their sub-token embeddings. Nevertheless, this method produced a performance inferior to that of generating multiple sub-tokens with ProMap$_G$.

\section{Conclusion} \label{sec:conclusion}
In this work, we introduced a new method dubbed ProMap for translating words between languages using multilingual pretrained language models. ProMap demonstrates strong performance in both rich-resource and low-resource languages. It is also able to achieve good results even with limited amounts of training data. Overall, we believe ProMap comprises an exciting advancement in bilingual lexicon induction and holds promise for improving translation in low-resource languages.

\section*{Limitations}

While the proposed ProMap model has demonstrated promising performance, it is important to highlight the following potential limitations:

\begin{itemize}
    \item The ProMap$_{G}$ model struggles to generate words of multiple sub-tokens, particularly when $n>1$. This limitation is primarily due to the complexity of word combinations that can be generated from multiple masked tokens. In cases of languages with rich morphology such as Arabic, this situation is even more challenging due to the vast number of possible combinations a word can have. 

    \item The performance of ProMap$_S$ heavily depends on the \textbf{P@K} performance achieved by the static WEs alignment method, and therefore, in the case of the few-shot learning, it is hard to achieve better results using this variant.
    
    \item Finetuning large PLMs is a time-consuming process, making the task of finding optimal hyperparameters labor-intensive. Additionally, finetuning large PLMs poses a significant challenge in reproducing results, requiring multiple runs to achieve consistent results.

    \item We evaluate our approach for multilingual scenarios using the XLM-17 mPLM, which currently supports $17$ languages. However, it should be noted that not all languages in the dictionaries dataset we used are covered by XLM-17. It is also worth experimenting with language models with larger vocabularies and fewer languages as a way to alleviate challenges compounded by the curse of multilinguality caused by mPLMs where per-language performance drops as with the increase of languages in the mPLM \cite{conneau-etal-2020-unsupervised}.


\end{itemize}

\section*{Ethics Statement}
This research aims to improve language technology for under-resourced languages by addressing lexical disparities between languages, groups, and cultures. The focus is on bilingual lexicon induction, a vital aspect in cross-lingual NLP with implications for machine translation and other tasks. The study includes various language families and all Arabic dialects, which are spoken by $\sim 450$M people. The goal is to expand NLP methods to lower-resource and under-represented languages using few-shot techniques. Ultimately, our work seeks to increase access to technology by serving diverse populations. 

The data used in our work, word translation pairs, is publicly accessible and in our view poses no risks. For any real-world use, we strongly suggest extensive evaluations and analyses be made before deployment. We also encourage use of our work in pro-social contexts such as health and language education.

\bibliography{anthology,custom}

\begin{thebibliography}{48}
\expandafter\ifx\csname natexlab\endcsname\relax\def\natexlab#1{#1}\fi

\bibitem[{Abdul-Mageed et~al.(2021)Abdul-Mageed, Elmadany, and
  Nagoudi}]{abdul-mageed-etal-2021-arbert}
Muhammad Abdul-Mageed, AbdelRahim Elmadany, and El~Moatez~Billah Nagoudi. 2021.
\newblock \href {https://doi.org/10.18653/v1/2021.acl-long.551} {{ARBERT} {\&}
  {MARBERT}: Deep bidirectional transformers for {A}rabic}.
\newblock In \emph{Proceedings of the 59th Annual Meeting of the Association
  for Computational Linguistics and the 11th International Joint Conference on
  Natural Language Processing (Volume 1: Long Papers)}, pages 7088--7105,
  Online. Association for Computational Linguistics.

\bibitem[{Artetxe et~al.(2016)Artetxe, Labaka, and
  Agirre}]{artetxe-etal-2016-learning}
Mikel Artetxe, Gorka Labaka, and Eneko Agirre. 2016.
\newblock \href {https://doi.org/10.18653/v1/D16-1250} {Learning principled
  bilingual mappings of word embeddings while preserving monolingual
  invariance}.
\newblock In \emph{Proceedings of the 2016 Conference on Empirical Methods in
  Natural Language Processing}, pages 2289--2294, Austin, Texas. Association
  for Computational Linguistics.

\bibitem[{Artetxe et~al.(2018{\natexlab{a}})Artetxe, Labaka, and
  Agirre}]{Artetxe_Labaka_Agirre_2018}
Mikel Artetxe, Gorka Labaka, and Eneko Agirre. 2018{\natexlab{a}}.
\newblock \href {https://doi.org/10.1609/aaai.v32i1.11992} {Generalizing and
  improving bilingual word embedding mappings with a multi-step framework of
  linear transformations}.
\newblock \emph{Proceedings of the AAAI Conference on Artificial Intelligence},
  32(1).

\bibitem[{Artetxe et~al.(2018{\natexlab{b}})Artetxe, Labaka, and
  Agirre}]{artetxe-etal-2018-robust}
Mikel Artetxe, Gorka Labaka, and Eneko Agirre. 2018{\natexlab{b}}.
\newblock \href {https://doi.org/10.18653/v1/P18-1073} {A robust self-learning
  method for fully unsupervised cross-lingual mappings of word embeddings}.
\newblock In \emph{Proceedings of the 56th Annual Meeting of the Association
  for Computational Linguistics (Volume 1: Long Papers)}, pages 789--798,
  Melbourne, Australia. Association for Computational Linguistics.

\bibitem[{Artetxe et~al.(2020)Artetxe, Ruder, Yogatama, Labaka, and
  Agirre}]{artetxe-etal-2020-call}
Mikel Artetxe, Sebastian Ruder, Dani Yogatama, Gorka Labaka, and Eneko Agirre.
  2020.
\newblock \href {https://doi.org/10.18653/v1/2020.acl-main.658} {A call for
  more rigor in unsupervised cross-lingual learning}.
\newblock In \emph{Proceedings of the 58th Annual Meeting of the Association
  for Computational Linguistics}, pages 7375--7388, Online. Association for
  Computational Linguistics.

\bibitem[{Bouamor et~al.(2018)Bouamor, Habash, Salameh, Zaghouani, Rambow,
  Abdulrahim, Obeid, Khalifa, Eryani, Erdmann, and
  Oflazer}]{bouamor-etal-2018-madar}
Houda Bouamor, Nizar Habash, Mohammad Salameh, Wajdi Zaghouani, Owen Rambow,
  Dana Abdulrahim, Ossama Obeid, Salam Khalifa, Fadhl Eryani, Alexander
  Erdmann, and Kemal Oflazer. 2018.
\newblock \href {https://aclanthology.org/L18-1535} {The {MADAR} {A}rabic
  dialect corpus and lexicon}.
\newblock In \emph{Proceedings of the Eleventh International Conference on
  Language Resources and Evaluation ({LREC} 2018)}, Miyazaki, Japan. European
  Language Resources Association (ELRA).

\bibitem[{Conneau et~al.(2020)Conneau, Khandelwal, Goyal, Chaudhary, Wenzek,
  Guzm{\'a}n, Grave, Ott, Zettlemoyer, and
  Stoyanov}]{conneau-etal-2020-unsupervised}
Alexis Conneau, Kartikay Khandelwal, Naman Goyal, Vishrav Chaudhary, Guillaume
  Wenzek, Francisco Guzm{\'a}n, Edouard Grave, Myle Ott, Luke Zettlemoyer, and
  Veselin Stoyanov. 2020.
\newblock \href {https://doi.org/10.18653/v1/2020.acl-main.747} {Unsupervised
  cross-lingual representation learning at scale}.
\newblock In \emph{Proceedings of the 58th Annual Meeting of the Association
  for Computational Linguistics}, pages 8440--8451, Online. Association for
  Computational Linguistics.

\bibitem[{Devlin et~al.(2019)Devlin, Chang, Lee, and
  Toutanova}]{devlin-etal-2019-bert}
Jacob Devlin, Ming-Wei Chang, Kenton Lee, and Kristina Toutanova. 2019.
\newblock \href {https://doi.org/10.18653/v1/N19-1423} {{BERT}: Pre-training of
  deep bidirectional transformers for language understanding}.
\newblock In \emph{Proceedings of the 2019 Conference of the North {A}merican
  Chapter of the Association for Computational Linguistics: Human Language
  Technologies, Volume 1 (Long and Short Papers)}, pages 4171--4186,
  Minneapolis, Minnesota. Association for Computational Linguistics.

\bibitem[{El~Mekki et~al.(2021)El~Mekki, El~Mahdaouy, Berrada, and
  Khoumsi}]{El2021On}
Abdellah El~Mekki, Abdelkader El~Mahdaouy, Ismail Berrada, and Ahmed Khoumsi.
  2021.
\newblock On the {Role} of {Orthographic} {Variations} in {Building}
  {Multidialectal} {Arabic} {Word} {Embeddings}.
\newblock \emph{Proceedings of the Canadian Conference on Artificial
  Intelligence}.
\newblock Https://caiac.pubpub.org/pub/pdf9jqoh.

\bibitem[{Erdmann et~al.(2018)Erdmann, Zalmout, and
  Habash}]{erdmann-etal-2018-addressing}
Alexander Erdmann, Nasser Zalmout, and Nizar Habash. 2018.
\newblock \href {https://doi.org/10.18653/v1/P18-2089} {Addressing noise in
  multidialectal word embeddings}.
\newblock In \emph{Proceedings of the 56th Annual Meeting of the Association
  for Computational Linguistics (Volume 2: Short Papers)}, pages 558--565,
  Melbourne, Australia. Association for Computational Linguistics.

\bibitem[{Feng et~al.(2022)Feng, Cao, Zhao, Wang, and
  Peng}]{feng-etal-2022-cross}
Zihao Feng, Hailong Cao, Tiejun Zhao, Weixuan Wang, and Wei Peng. 2022.
\newblock \href {https://aclanthology.org/2022.coling-1.469} {Cross-lingual
  feature extraction from monolingual corpora for low-resource unsupervised
  bilingual lexicon induction}.
\newblock In \emph{Proceedings of the 29th International Conference on
  Computational Linguistics}, pages 5278--5287, Gyeongju, Republic of Korea.
  International Committee on Computational Linguistics.

\bibitem[{Gao et~al.(2021)Gao, Fisch, and Chen}]{gao-etal-2021-making}
Tianyu Gao, Adam Fisch, and Danqi Chen. 2021.
\newblock \href {https://doi.org/10.18653/v1/2021.acl-long.295} {Making
  pre-trained language models better few-shot learners}.
\newblock In \emph{Proceedings of the 59th Annual Meeting of the Association
  for Computational Linguistics and the 11th International Joint Conference on
  Natural Language Processing (Volume 1: Long Papers)}, pages 3816--3830,
  Online. Association for Computational Linguistics.

\bibitem[{Glava{\v{s}} et~al.(2019)Glava{\v{s}}, Litschko, Ruder, and
  Vuli{\'c}}]{glavas-etal-2019-properly}
Goran Glava{\v{s}}, Robert Litschko, Sebastian Ruder, and Ivan Vuli{\'c}. 2019.
\newblock \href {https://doi.org/10.18653/v1/P19-1070} {How to (properly)
  evaluate cross-lingual word embeddings: On strong baselines, comparative
  analyses, and some misconceptions}.
\newblock In \emph{Proceedings of the 57th Annual Meeting of the Association
  for Computational Linguistics}, pages 710--721, Florence, Italy. Association
  for Computational Linguistics.

\bibitem[{Glava{\v{s}} and Vuli{\'c}(2020)}]{glavas-vulic-2020-non}
Goran Glava{\v{s}} and Ivan Vuli{\'c}. 2020.
\newblock \href {https://doi.org/10.18653/v1/2020.acl-main.675} {Non-linear
  instance-based cross-lingual mapping for non-isomorphic embedding spaces}.
\newblock In \emph{Proceedings of the 58th Annual Meeting of the Association
  for Computational Linguistics}, pages 7548--7555, Online. Association for
  Computational Linguistics.

\bibitem[{Habash et~al.(2018)Habash, Eryani, Khalifa, Rambow, Abdulrahim,
  Erdmann, Faraj, Zaghouani, Bouamor, Zalmout, Hassan, Al-Shargi, Alkhereyf,
  Abdulkareem, Eskander, Salameh, and Saddiki}]{habash-etal-2018-unified}
Nizar Habash, Fadhl Eryani, Salam Khalifa, Owen Rambow, Dana Abdulrahim,
  Alexander Erdmann, Reem Faraj, Wajdi Zaghouani, Houda Bouamor, Nasser
  Zalmout, Sara Hassan, Faisal Al-Shargi, Sakhar Alkhereyf, Basma Abdulkareem,
  Ramy Eskander, Mohammad Salameh, and Hind Saddiki. 2018.
\newblock \href {https://aclanthology.org/L18-1574} {Unified guidelines and
  resources for {A}rabic dialect orthography}.
\newblock In \emph{Proceedings of the Eleventh International Conference on
  Language Resources and Evaluation ({LREC} 2018)}, Miyazaki, Japan. European
  Language Resources Association (ELRA).

\bibitem[{Inoue et~al.(2021)Inoue, Alhafni, Baimukan, Bouamor, and
  Habash}]{inoue-etal-2021-interplay}
Go~Inoue, Bashar Alhafni, Nurpeiis Baimukan, Houda Bouamor, and Nizar Habash.
  2021.
\newblock \href {https://aclanthology.org/2021.wanlp-1.10} {The interplay of
  variant, size, and task type in {A}rabic pre-trained language models}.
\newblock In \emph{Proceedings of the Sixth Arabic Natural Language Processing
  Workshop}, pages 92--104, Kyiv, Ukraine (Virtual). Association for
  Computational Linguistics.

\bibitem[{Jawanpuria et~al.(2019)Jawanpuria, Balgovind, Kunchukuttan, and
  Mishra}]{jawanpuria-etal-2019-learning}
Pratik Jawanpuria, Arjun Balgovind, Anoop Kunchukuttan, and Bamdev Mishra.
  2019.
\newblock \href {https://doi.org/10.1162/tacl_a_00257} {Learning multilingual
  word embeddings in latent metric space: A geometric approach}.
\newblock \emph{Transactions of the Association for Computational Linguistics},
  7:107--120.

\bibitem[{Joulin et~al.(2018)Joulin, Bojanowski, Mikolov, J{\'e}gou, and
  Grave}]{joulin-etal-2018-loss}
Armand Joulin, Piotr Bojanowski, Tomas Mikolov, Herv{\'e} J{\'e}gou, and
  Edouard Grave. 2018.
\newblock \href {https://doi.org/10.18653/v1/D18-1330} {Loss in translation:
  Learning bilingual word mapping with a retrieval criterion}.
\newblock In \emph{Proceedings of the 2018 Conference on Empirical Methods in
  Natural Language Processing}, pages 2979--2984, Brussels, Belgium.
  Association for Computational Linguistics.

\bibitem[{Karan et~al.(2020)Karan, Vuli{\'c}, Korhonen, and
  Glava{\v{s}}}]{karan-etal-2020-classification}
Mladen Karan, Ivan Vuli{\'c}, Anna Korhonen, and Goran Glava{\v{s}}. 2020.
\newblock \href {https://doi.org/10.18653/v1/2020.acl-main.618}
  {Classification-based self-learning for weakly supervised bilingual lexicon
  induction}.
\newblock In \emph{Proceedings of the 58th Annual Meeting of the Association
  for Computational Linguistics}, pages 6915--6922, Online. Association for
  Computational Linguistics.

\bibitem[{Lample and Conneau(2019)}]{lample2019cross}
Guillaume Lample and Alexis Conneau. 2019.
\newblock Cross-lingual language model pretraining.
\newblock \emph{Advances in Neural Information Processing Systems (NeurIPS)}.

\bibitem[{Lample et~al.(2018)Lample, Conneau, Ranzato, Denoyer, and
  Jégou}]{lample2018word}
Guillaume Lample, Alexis Conneau, Marc'Aurelio Ranzato, Ludovic Denoyer, and
  Hervé Jégou. 2018.
\newblock \href {https://openreview.net/forum?id=H196sainb} {Word translation
  without parallel data}.
\newblock In \emph{International Conference on Learning Representations}.

\bibitem[{Lan et~al.(2020)Lan, Chen, Xu, and Ritter}]{lan-etal-2020-empirical}
Wuwei Lan, Yang Chen, Wei Xu, and Alan Ritter. 2020.
\newblock \href {https://doi.org/10.18653/v1/2020.emnlp-main.382} {An empirical
  study of pre-trained transformers for {A}rabic information extraction}.
\newblock In \emph{Proceedings of the 2020 Conference on Empirical Methods in
  Natural Language Processing (EMNLP)}, pages 4727--4734, Online. Association
  for Computational Linguistics.

\bibitem[{Li et~al.(2022)Li, Liu, Collier, Korhonen, and
  Vuli{\'c}}]{li-etal-2022-improving}
Yaoyiran Li, Fangyu Liu, Nigel Collier, Anna Korhonen, and Ivan Vuli{\'c}.
  2022.
\newblock \href {https://doi.org/10.18653/v1/2022.acl-long.299} {Improving word
  translation via two-stage contrastive learning}.
\newblock In \emph{Proceedings of the 60th Annual Meeting of the Association
  for Computational Linguistics (Volume 1: Long Papers)}, pages 4353--4374,
  Dublin, Ireland. Association for Computational Linguistics.

\bibitem[{Mallinson et~al.(2020)Mallinson, Severyn, Malmi, and
  Garrido}]{mallinson-etal-2020-felix}
Jonathan Mallinson, Aliaksei Severyn, Eric Malmi, and Guillermo Garrido. 2020.
\newblock \href {https://doi.org/10.18653/v1/2020.findings-emnlp.111} {{FELIX}:
  Flexible text editing through tagging and insertion}.
\newblock In \emph{Findings of the Association for Computational Linguistics:
  EMNLP 2020}, pages 1244--1255, Online. Association for Computational
  Linguistics.

\bibitem[{Malmi et~al.(2020)Malmi, Severyn, and
  Rothe}]{malmi-etal-2020-unsupervised}
Eric Malmi, Aliaksei Severyn, and Sascha Rothe. 2020.
\newblock \href {https://doi.org/10.18653/v1/2020.emnlp-main.699} {Unsupervised
  text style transfer with padded masked language models}.
\newblock In \emph{Proceedings of the 2020 Conference on Empirical Methods in
  Natural Language Processing (EMNLP)}, pages 8671--8680, Online. Association
  for Computational Linguistics.

\bibitem[{Marchisio et~al.(2020)Marchisio, Duh, and
  Koehn}]{marchisio-etal-2020-unsupervised}
Kelly Marchisio, Kevin Duh, and Philipp Koehn. 2020.
\newblock \href {https://aclanthology.org/2020.wmt-1.68} {When does
  unsupervised machine translation work?}
\newblock In \emph{Proceedings of the Fifth Conference on Machine Translation},
  pages 571--583, Online. Association for Computational Linguistics.

\bibitem[{Marchisio et~al.(2022)Marchisio, Saad-Eldin, Duh, Priebe, and
  Koehn}]{graph-bli-low}
Kelly Marchisio, Ali Saad-Eldin, Kevin Duh, Carey Priebe, and Philipp Koehn.
  2022.
\newblock \href {https://doi.org/10.48550/ARXIV.2210.14378} {Bilingual lexicon
  induction for low-resource languages using graph matching via optimal
  transport}.

\bibitem[{Mikolov et~al.(2013)Mikolov, Le, and
  Sutskever}]{mikolov2013exploiting}
Tom{\'{a}}s Mikolov, Quoc~V. Le, and Ilya Sutskever. 2013.
\newblock \href {http://arxiv.org/abs/1309.4168} {Exploiting similarities among
  languages for machine translation}.
\newblock \emph{CoRR}, abs/1309.4168.

\bibitem[{Mohiuddin et~al.(2020)Mohiuddin, Bari, and
  Joty}]{mohiuddin-etal-2020-lnmap}
Tasnim Mohiuddin, M~Saiful Bari, and Shafiq Joty. 2020.
\newblock \href {https://doi.org/10.18653/v1/2020.emnlp-main.215} {{LNM}ap:
  Departures from isomorphic assumption in bilingual lexicon induction through
  non-linear mapping in latent space}.
\newblock In \emph{Proceedings of the 2020 Conference on Empirical Methods in
  Natural Language Processing (EMNLP)}, pages 2712--2723, Online. Association
  for Computational Linguistics.

\bibitem[{Nakashole and Flauger(2018)}]{nakashole-flauger-2018-characterizing}
Ndapa Nakashole and Raphael Flauger. 2018.
\newblock \href {https://doi.org/10.18653/v1/P18-2036} {Characterizing
  departures from linearity in word translation}.
\newblock In \emph{Proceedings of the 56th Annual Meeting of the Association
  for Computational Linguistics (Volume 2: Short Papers)}, pages 221--227,
  Melbourne, Australia. Association for Computational Linguistics.

\bibitem[{Ormazabal et~al.(2019)Ormazabal, Artetxe, Labaka, Soroa, and
  Agirre}]{ormazabal-etal-2019-analyzing}
Aitor Ormazabal, Mikel Artetxe, Gorka Labaka, Aitor Soroa, and Eneko Agirre.
  2019.
\newblock \href {https://doi.org/10.18653/v1/P19-1492} {Analyzing the
  limitations of cross-lingual word embedding mappings}.
\newblock In \emph{Proceedings of the 57th Annual Meeting of the Association
  for Computational Linguistics}, pages 4990--4995, Florence, Italy.
  Association for Computational Linguistics.

\bibitem[{Patra et~al.(2019)Patra, Moniz, Garg, Gormley, and
  Neubig}]{patra-etal-2019-bilingual}
Barun Patra, Joel Ruben~Antony Moniz, Sarthak Garg, Matthew~R. Gormley, and
  Graham Neubig. 2019.
\newblock \href {https://doi.org/10.18653/v1/P19-1018} {Bilingual lexicon
  induction with semi-supervision in non-isometric embedding spaces}.
\newblock In \emph{Proceedings of the 57th Annual Meeting of the Association
  for Computational Linguistics}, pages 184--193, Florence, Italy. Association
  for Computational Linguistics.

\bibitem[{Riley and Gildea(2018)}]{riley-gildea-2018-orthographic}
Parker Riley and Daniel Gildea. 2018.
\newblock \href {https://doi.org/10.18653/v1/P18-2062} {Orthographic features
  for bilingual lexicon induction}.
\newblock In \emph{Proceedings of the 56th Annual Meeting of the Association
  for Computational Linguistics (Volume 2: Short Papers)}, pages 390--394,
  Melbourne, Australia. Association for Computational Linguistics.

\bibitem[{Sachidananda et~al.(2021)Sachidananda, Yang, and
  Zhu}]{sachidananda2021filtered}
Vin Sachidananda, Ziyi Yang, and Chenguang Zhu. 2021.
\newblock \href {https://openreview.net/forum?id=A2gNouoXE7} {Filtered inner
  product projection for crosslingual embedding alignment}.
\newblock In \emph{International Conference on Learning Representations}.

\bibitem[{Salloum and Habash(2014)}]{SALLOUM2014372}
Wael Salloum and Nizar Habash. 2014.
\newblock \href {https://doi.org/https://doi.org/10.1016/j.jksuci.2014.06.010}
  {Adam: Analyzer for dialectal arabic morphology}.
\newblock \emph{Journal of King Saud University - Computer and Information
  Sciences}, 26(4):372--378.
\newblock Special Issue on Arabic NLP.

\bibitem[{Schick et~al.(2020)Schick, Schmid, and
  Sch{\"u}tze}]{schick-etal-2020-automatically}
Timo Schick, Helmut Schmid, and Hinrich Sch{\"u}tze. 2020.
\newblock \href {https://doi.org/10.18653/v1/2020.coling-main.488}
  {Automatically identifying words that can serve as labels for few-shot text
  classification}.
\newblock In \emph{Proceedings of the 28th International Conference on
  Computational Linguistics}, pages 5569--5578, Barcelona, Spain (Online).
  International Committee on Computational Linguistics.

\bibitem[{Schick and
  Sch{\"u}tze(2021{\natexlab{a}})}]{schick-schutze-2021-exploiting}
Timo Schick and Hinrich Sch{\"u}tze. 2021{\natexlab{a}}.
\newblock \href {https://doi.org/10.18653/v1/2021.eacl-main.20} {Exploiting
  cloze-questions for few-shot text classification and natural language
  inference}.
\newblock In \emph{Proceedings of the 16th Conference of the European Chapter
  of the Association for Computational Linguistics: Main Volume}, pages
  255--269, Online. Association for Computational Linguistics.

\bibitem[{Schick and
  Sch{\"u}tze(2021{\natexlab{b}})}]{schick-schutze-2021-just}
Timo Schick and Hinrich Sch{\"u}tze. 2021{\natexlab{b}}.
\newblock \href {https://doi.org/10.18653/v1/2021.naacl-main.185} {It{'}s not
  just size that matters: Small language models are also few-shot learners}.
\newblock In \emph{Proceedings of the 2021 Conference of the North American
  Chapter of the Association for Computational Linguistics: Human Language
  Technologies}, pages 2339--2352, Online. Association for Computational
  Linguistics.

\bibitem[{Shi et~al.(2021)Shi, Zettlemoyer, and Wang}]{shi-etal-2021-bilingual}
Haoyue Shi, Luke Zettlemoyer, and Sida~I. Wang. 2021.
\newblock \href {https://doi.org/10.18653/v1/2021.acl-long.67} {Bilingual
  lexicon induction via unsupervised bitext construction and word alignment}.
\newblock In \emph{Proceedings of the 59th Annual Meeting of the Association
  for Computational Linguistics and the 11th International Joint Conference on
  Natural Language Processing (Volume 1: Long Papers)}, pages 813--826, Online.
  Association for Computational Linguistics.

\bibitem[{Shin et~al.(2020)Shin, Razeghi, Logan~IV, Wallace, and
  Singh}]{shin-etal-2020-autoprompt}
Taylor Shin, Yasaman Razeghi, Robert~L. Logan~IV, Eric Wallace, and Sameer
  Singh. 2020.
\newblock \href {https://doi.org/10.18653/v1/2020.emnlp-main.346}
  {{A}uto{P}rompt: {E}liciting {K}nowledge from {L}anguage {M}odels with
  {A}utomatically {G}enerated {P}rompts}.
\newblock In \emph{Proceedings of the 2020 Conference on Empirical Methods in
  Natural Language Processing (EMNLP)}, pages 4222--4235, Online. Association
  for Computational Linguistics.

\bibitem[{Smith et~al.(2017)Smith, Turban, Hamblin, and
  Hammerla}]{smith2017offline}
Samuel~L. Smith, David H.~P. Turban, Steven Hamblin, and Nils~Y. Hammerla.
  2017.
\newblock \href {https://openreview.net/forum?id=r1Aab85gg} {Offline bilingual
  word vectors, orthogonal transformations and the inverted softmax}.
\newblock In \emph{International Conference on Learning Representations}.

\bibitem[{S{\o}gaard et~al.(2018)S{\o}gaard, Ruder, and
  Vuli{\'c}}]{sogaard-etal-2018-limitations}
Anders S{\o}gaard, Sebastian Ruder, and Ivan Vuli{\'c}. 2018.
\newblock \href {https://doi.org/10.18653/v1/P18-1072} {On the limitations of
  unsupervised bilingual dictionary induction}.
\newblock In \emph{Proceedings of the 56th Annual Meeting of the Association
  for Computational Linguistics (Volume 1: Long Papers)}, pages 778--788,
  Melbourne, Australia. Association for Computational Linguistics.

\bibitem[{van~der Maaten and Hinton(2008)}]{JMLR:v9:vandermaaten08a}
Laurens van~der Maaten and Geoffrey Hinton. 2008.
\newblock \href {http://jmlr.org/papers/v9/vandermaaten08a.html} {Visualizing
  data using t-sne}.
\newblock \emph{Journal of Machine Learning Research}, 9(86):2579--2605.

\bibitem[{Vuli{\'c} et~al.(2019)Vuli{\'c}, Glava{\v{s}}, Reichart, and
  Korhonen}]{vulic-etal-2019-really}
Ivan Vuli{\'c}, Goran Glava{\v{s}}, Roi Reichart, and Anna Korhonen. 2019.
\newblock \href {https://doi.org/10.18653/v1/D19-1449} {Do we really need fully
  unsupervised cross-lingual embeddings?}
\newblock In \emph{Proceedings of the 2019 Conference on Empirical Methods in
  Natural Language Processing and the 9th International Joint Conference on
  Natural Language Processing (EMNLP-IJCNLP)}, pages 4407--4418, Hong Kong,
  China. Association for Computational Linguistics.

\bibitem[{Xing et~al.(2015)Xing, Wang, Liu, and
  Lin}]{xing-etal-2015-normalized}
Chao Xing, Dong Wang, Chao Liu, and Yiye Lin. 2015.
\newblock \href {https://doi.org/10.3115/v1/N15-1104} {Normalized word
  embedding and orthogonal transform for bilingual word translation}.
\newblock In \emph{Proceedings of the 2015 Conference of the North {A}merican
  Chapter of the Association for Computational Linguistics: Human Language
  Technologies}, pages 1006--1011, Denver, Colorado. Association for
  Computational Linguistics.

\bibitem[{Zhang et~al.(2021)Zhang, Ji, Xiao, Duan, Zhang, Shi, and
  Luo}]{zhang-etal-2021-combining}
Jinpeng Zhang, Baijun Ji, Nini Xiao, Xiangyu Duan, Min Zhang, Yangbin Shi, and
  Weihua Luo. 2021.
\newblock \href {https://doi.org/10.18653/v1/2021.findings-acl.260} {Combining
  static word embeddings and contextual representations for bilingual lexicon
  induction}.
\newblock In \emph{Findings of the Association for Computational Linguistics:
  ACL-IJCNLP 2021}, pages 2943--2955, Online. Association for Computational
  Linguistics.

\bibitem[{Zhang et~al.(2017{\natexlab{a}})Zhang, Liu, Luan, and
  Sun}]{zhang-etal-2017-adversarial}
Meng Zhang, Yang Liu, Huanbo Luan, and Maosong Sun. 2017{\natexlab{a}}.
\newblock \href {https://doi.org/10.18653/v1/P17-1179} {Adversarial training
  for unsupervised bilingual lexicon induction}.
\newblock In \emph{Proceedings of the 55th Annual Meeting of the Association
  for Computational Linguistics (Volume 1: Long Papers)}, pages 1959--1970,
  Vancouver, Canada. Association for Computational Linguistics.

\bibitem[{Zhang et~al.(2017{\natexlab{b}})Zhang, Liu, Luan, and
  Sun}]{zhang-etal-2017-earth}
Meng Zhang, Yang Liu, Huanbo Luan, and Maosong Sun. 2017{\natexlab{b}}.
\newblock \href {https://doi.org/10.18653/v1/D17-1207} {Earth mover{'}s
  distance minimization for unsupervised bilingual lexicon induction}.
\newblock In \emph{Proceedings of the 2017 Conference on Empirical Methods in
  Natural Language Processing}, pages 1934--1945, Copenhagen, Denmark.
  Association for Computational Linguistics.

\end{thebibliography}
\bibliographystyle{acl_natbib}

\clearpage

\appendix

\section*{Appendices}
\noindent We provide an overview of the Appendix below. \\

\begin{enumerate}[I]
\setlength\itemsep{1.3em}

 \item \noindent\textbf{Method (Appendix~\ref{sec:method-appendix}).} 
 
 \par This section  provides more details about the prompting-based finetuning of BERT-like PLMs.
        
\item \noindent\textbf{Experiment Settings (Appendix~\ref{sec:experiments-settings-app}).}

 \par This section gives additional information about the considered experiment environments.
        \begin{itemize}
            \item We provide implementation details in Appendix~\ref{sec:implementation-details-app}.
            \item  We describe the computing infrastructure in Appendix~\ref{sec:computing-infrastructure-app}.
            \item We show the average runtimes of the competitive approaches  in Appendix~\ref{sec:average-runtimes-app}.
            \item We give more insights about the MADAR lexicon in Appendix ~\ref{sec:madar-insights}.
            \item We provide the links of the used datasets in ~\ref{sec:datasets-links-app}.
        \end{itemize}

\item \noindent\textbf{Baseline Systems (Appendix~\ref{sec:appendix-baselines}).}

 \par   This section presents the various baseline systems against which we compared our results.

\item \noindent\textbf{Analyses (Appendix~\ref{sec:analyses-appendix}).} 

\par Finally, we provide additional experiments and results, including:

       \begin{itemize}
            \item Examples of word translations by ProMap in ~\ref{sec:predictions-examples}.
            \item  Comparison between the performance of ProMap$_G$ on Dialectal Arabic using two competitive PLMs in Appendix~\ref{sec:plms-compare-app}.
            \item Comparison of the performance achieved by different prompt templates in Appendix~\ref{sec:effect-prompt-template}.
            \item The full results on all the Arabic pairs covered by the Madar Lexicon in Appendix~\ref{sec:madar-full-results}.
            \item The full results of the few-shot experiments on the multilingual scenario in Appendix~\ref{sec:few-shot-res}.
            \item Various t-SNE visualizations of the WEs before and after finetuning in Appendix~\ref{sec:visu-tsne}.
        \end{itemize}

\end{enumerate}


\section{Prompt-based Finetuning Method}
\label{sec:method-appendix}
Prompt-based finetuning involves using natural language templates to represent input statements and treating text classification tasks as cloze-style tasks. For example, in sentence classification, if we need to classify the sentence "The Moroccan team made it to the world cup semi-final" as either  $y_1 = POLITICS$ or $y_2 = SPORT$, the template might look like this:

$$
x_p  = \left[ {{\rm{CLS}}} \right]x{\rm{, a}}\left[ {{\rm{MASK}}} \right]{\rm{topic}}
$$

With $x = \left\{ {x_1 ,...,x_l } \right\}$ is the input sentence of $l$ tokens. Using masked language modeling as the finetuning task, the $[MASK]$ token in $x_p$ will have a predicted token $v_p \in V  = \left\{ {v_1 ,...,v_r } \right\}$ where $V$ is the vocabulary covered by the PLM with size $r$. Then, the value $v_p$ should be mapped to the final label (i.e. POLITICS or SPORT for the example of sentence $x$). The objective is to extract the token $v_p$ from $V$ that can have the maximal probability to be filled in $[MASK]$. It can be noted as $ p\left( {\left[ {{\rm{MASK}}} \right] = v_p \in V |x_p } \right)$. To finetune a PLM using this prompt-based method for a classification task, all input sentences should first be designed as a unique template (such as $x_p$) when the ground truth label is replaced by the masked token $[MASK]$, and then train the model to infill the masked token with the class-token.

\section{Experiments}

\subsection{Settings}
\label{sec:experiments-settings-app}
\subsubsection{Implementation Details}
\label{sec:implementation-details-app}
Table \ref{tab:trainable-parameters} presents the number of trainable parameters for each mPLM used in our paper.

\begin{table}[!htb]
\centering
\resizebox{\columnwidth}{!}{%
\begin{tabular}{lc}
\toprule
Model                    & \# of trainable parameters \\ \midrule
\textbf{XLM-MLM-17-1280} & 571,696,960                      \\
\textbf{MARBERT}         & 162,942,880                      \\ \bottomrule
\end{tabular}}
\caption{The number of trainable parameters for each mPLM used for ProMap.}
\label{tab:trainable-parameters}
\end{table}

\subsubsection{Computing Infrastructure}
\label{sec:computing-infrastructure-app}
We conducted our experiments utilizing a workstation equipped with an Intel(R) Xeon(R) Silver 4216 CPU operating at 2.10GHz and a single Nvidia Tesla V100 GPU with 32GB of RAM.

\subsubsection{Average Runtimes}
\label{sec:average-runtimes-app}

\begin{table}[!htb]
\centering 

\begin{tabular}{ll}
\toprule
Method          & Average runtime \\ \midrule
\textbf{CLC1}   & 21m38s          \\
\textbf{ProMap$_G$} & 6m46s           \\
\textbf{ProMap$_S$} & 8m36s           \\ \bottomrule
\end{tabular}
\caption{Average training runtime of the ProMap and CLC1 methods. The runtime of ProMap$_S$ includes both the finetuning of ProMap and the re-ranking.}
\label{tab:average-runtimes}
\end{table}

\subsubsection{Insights about the Dialectal Arabic to MSA Data}
\label{sec:madar-insights}

To construct the dictionaries for word translation between Dialectal Arabic and MSA experiments, we utilized the MADAR lexicon, which encompasses 25 Arab cities. This lexicon provides an MSA translation for every Dialectal Arabic word. We grouped the words from cities within the same country to create a country-level dictionary. This resulted in dictionaries of 10 Arab countries. We then performed a random split to divide the data into training and testing sets. Table \ref{tab:data-sizes-madar} presents the train and test sizes for each country-level Arabic dialect to MSA dictionary.

\begin{table}[!htb]
\centering
\resizebox{\columnwidth}{!}{%
\begin{tabular}{lcc}
\toprule
\textbf{Arabic variants pairs}  & \textbf{\# of training pairs} & \textbf{\# of testing pairs} \\ \midrule
\textbf{Moroccan (MAR) $\to$ MSA} & 740                           & 193                          \\
\textbf{Algerian (ALG) $\to$ MSA} & 638                           & 161                          \\
\textbf{Tunisian (TUN) $\to$ MSA}   & 844                           & 209                          \\
\textbf{Libyan (LBY) $\to$ MSA}  & 879                           & 217                          \\
\textbf{Egyptian (EGY) $\to$ MSA}  & 1,077                          & 282                          \\
\textbf{Sudanese $\to$ MSA} & 1,322                          & 341                          \\
\textbf{Leventine (LEV) $\to$ MSA} & 1,111                          & 298                          \\
\textbf{Iraqi (IRQ) $\to$ MSA}  & 1,027                          & 255                          \\
\textbf{Gulf (GLF) $\to$ MSA} & 2,051                          & 526                          \\
\textbf{Yemeni (YEM) $\to$ MSA}  & 1,466                          & 350                          \\ \bottomrule
\end{tabular}}
\caption{Sizes of train and test datasets constructed from the MADAR lexicon for the case of country-level dialectal Arabic to MSA word translation.}
\label{tab:data-sizes-madar}
\end{table}

\subsubsection{Datasets Links}
\label{sec:datasets-links-app}
\paragraph{Multilingual Scenario:}
\begin{itemize}
    \item \href{https://github.com/codogogo/xling-eval}{XLING bilingual dictionaries} 
\end{itemize}

\paragraph{Multi-dialectal Scenario:}
\begin{itemize}
    \item \href{https://sites.google.com/nyu.edu/madar/}{MADAR lexicon} 
    \item \href{https://camel.abudhabi.nyu.edu/arabic-multidialectal-embeddings}{Arabic dialect to Arabic dialect lexicon} 
\end{itemize}

\subsection{Baseline Systems}
\label{sec:appendix-baselines}

In the first scenario, when evaluating the multilingual setting, we compare the performance of ProMap variants to the following baseline systems:

\begin{itemize}
    \item \textbf{RCSLS} \cite{joulin-etal-2018-loss} optimizes a convex relaxation of CSLS loss during training, and therefore it learns a non-orthogonal mapping and improves the supervised BLI performance.
    \item \textbf{Vecmap} \cite{Artetxe_Labaka_Agirre_2018} follows multiple steps to perform word translation between two languages. The steps are whitening, orthogonal mapping, re-weighting, de-whitening, and dimensionality reduction.
    \item \textbf{LNMap} \cite{mohiuddin-etal-2020-lnmap} uses non-linear autoencoders to learn a non-linear mapping of the static WEs of two languages into two latent spaces. It then uses these latent spaces to learn another non-linear mapping between them.
    \item \textbf{FIPP} \cite{sachidananda2021filtered} finds the common geometric structure between both languages' embeddings, then using the common structure, it aligns the Gram matrices of these embeddings.
    \item \textbf{CLC1} \cite{li-etal-2022-improving} refines the linear Vecmap framework via CL objective iterations.
    \item \textbf{CLC2} \cite{li-etal-2022-improving} combines the embeddings generated by CL1 and a multilingual PLM (optimized using a contrastive learning objective on the seed dictionary) aligned to the CLC1 embeddings. 
\end{itemize}

For the word translation between Arabic variants, we compare our results to the following approaches that have demonstrated good performance on the same task:

\begin{itemize}
    \item \textbf{\citet{erdmann-etal-2018-addressing}} presents the first version of the Vecamp framework, which uses a linear mapping to align the static word embeddings (WEs) of two languages, $L1$ and $L2$. This method employs the orthogonal Procrustes problem to learn the mapping.
    \item \textbf{\citet{artetxe-etal-2016-learning}} uses the same Vecmap version as \cite{erdmann-etal-2018-addressing} to align the static WEs of $L1$ and $L2$. In addition, it uses self-training iterations to allow the model to learn from a larger dictionary at each iteration.
    \item \textbf{\citet{riley-gildea-2018-orthographic}} extends the static WEs of $L1$ and $L2$ by incorporating orthographic features of the covered words. The Vecmap mapping is then applied to these extended WEs.
    \item \textbf{\citet{El2021On}} uses Canonical Correlation Analysis (CCA) to align the orthographic features in a shared space before extending the static WEs, as in \cite{riley-gildea-2018-orthographic}.
\end{itemize}

\section{Analyses}
\label{sec:analyses-appendix}
\subsection{Examples of Translations by ProMap}
\label{sec:predictions-examples}

Table \ref{tab:predictions-examples-multilingual} presents examples of translations predicted by ProMap variants and CLC1 for various language pairs. The table illustrates both instances when ProMap variants accurately predict translations and instances when it fails. Additionally, the table displays the sub-tokens generated by the ProMap$_G$ variant. As demonstrated by the provided examples, ProMap variants are capable to predict correct translations, even for distant languages such as Turkish-Italian, where both ProMap variants were able to correctly predict translations while the CLC1 model failed. Additionally, there are cases where only ProMap$_G$ predicts the correct translations even if it contains more than one sub-token. This indicates that the non-autoregressive word translation method for the mPLM can independently generate correct sub-tokens that form the correct word translation. Furthermore, ProMap$_S$ demonstrated in some cases to be the only successful model, highlighting the power of the re-ranking mechanism implemented in our approach.

In the same vein, Table \ref{tab:arabic_examples} presents examples of predictions generated ProMap$_{G}$ applied on MARBERT. These examples demonstrate the ability of this model to handle word translation between different Arabic dialects and MSA. Also, the table illustrates that in most cases, the correct predictions can be found within the top-5 predictions. However, the model appears to have difficulties in translating from MSA to dialectal Arabic in some instances, in contrast to the translation from dialectal Arabic to MSA which is accurate in the majority of examples.

\begin{table*}[!htb]
\centering
\resizebox{2.\columnwidth}{!}{%
\begin{tabular}{lll|llll}
\toprule
\textbf{Pair}  & \textbf{Source Word}  & \textbf{True Translation} & \textbf{CLC1}                       & \textbf{ProMap$_G$ sub-tokens}                 & \textbf{ProMap$_G$}                                  & \textbf{ProMap$_S$}                       \\ \midrule
\textbf{DE-FR} & animationen           & animations                & \cellcolor[HTML]{FFCCC9}animées       & anim, ations, {[}PAD{]}, {[}PAD{]}         & \cellcolor[HTML]{9AFF99}animations                & \cellcolor[HTML]{9AFF99}animations     \\
\textbf{DE-FR} & infinitesimalrechnung & calcul                    & \cellcolor[HTML]{FFCCC9}infinitésimal & calcul, {[}PAD{]}, {[}PAD{]}, {[}PAD{]}    & \cellcolor[HTML]{9AFF99}calcul                    & \cellcolor[HTML]{FFCCC9}infinitésimal  \\
\textbf{DE-FR} & erniedrigung          & humiliation               & \cellcolor[HTML]{FFCCC9}privation     & humili, ation, {[}PAD{]}, {[}PAD{]}        & \cellcolor[HTML]{9AFF99}humiliation               & \cellcolor[HTML]{9AFF99}humiliation    \\
\textbf{EN-IT} & grille                & griglia                   & \cellcolor[HTML]{FFCCC9}calandra      & gr, iglia, {[}PAD{]}, {[}PAD{]}            & \cellcolor[HTML]{9AFF99}griglia                   & \cellcolor[HTML]{FFCCC9}calandra       \\
\textbf{EN-IT} & selector              & selettore                 & \cellcolor[HTML]{FFCCC9}selezionatore & selet, tore, {[}PAD{]}, {[}PAD{]}          & \cellcolor[HTML]{9AFF99}selettore                 & \cellcolor[HTML]{9AFF99}selettore      \\
\textbf{EN-IT} & consulate             & consolato                 & \cellcolor[HTML]{FFCCC9}ambasciata    & consul, ato, {[}PAD{]}, {[}PAD{]}          & \cellcolor[HTML]{FFCCC9}consulato                 & \cellcolor[HTML]{9AFF99}consolato      \\
\textbf{TR-IT} & hatırlatır            & ricorda                   & \cellcolor[HTML]{FFCCC9}rammenta      & ricorda, {[}PAD{]}, {[}PAD{]}, {[}PAD{]}   & \cellcolor[HTML]{9AFF99}ricorda                   & \cellcolor[HTML]{9AFF99}ricorda        \\
\textbf{TR-IT} & gezi                  & escursione                & \cellcolor[HTML]{FFCCC9}passeggiata   & escur, aggio, {[}PAD{]}, {[}PAD{]}         & \cellcolor[HTML]{FFCCC9}escuraggio                & \cellcolor[HTML]{9AFF99}escursione     \\
\textbf{TR-IT} & fosforilasyon         & fosforilazione            & \cellcolor[HTML]{FFCCC9}pathway       & fosfor, dazione, {[}PAD{]}, {[}PAD{]}      & \cellcolor[HTML]{FFCCC9}fosfordazione             & \cellcolor[HTML]{9AFF99}fosforilazione \\
\textbf{TR-IT} & aldatma               & inganno                   & \cellcolor[HTML]{9AFF99}inganno       & donazione, {[}PAD{]}, {[}PAD{]}, {[}PAD{]} & \cellcolor[HTML]{FFCCC9}donazione                 & \cellcolor[HTML]{FFCCC9}seduzione      \\
\textbf{EN-FR} & abbreviation          & abréviation               & \cellcolor[HTML]{9AFF99}abréviation   & sigle, {[}PAD{]}, {[}PAD{]}, {[}PAD{]}     & \cellcolor[HTML]{FFCCC9}sigle                     & \cellcolor[HTML]{FFCCC9}sigle          \\
\textbf{EN-FR} & presumed              & présumé                   & \cellcolor[HTML]{9AFF99}présumé       & sup, posé, {[}PAD{]}, {[}PAD{]}            & \cellcolor[HTML]{FFCCC9}supposé                   & \cellcolor[HTML]{FFCCC9}supposé        \\
\textbf{TR-FR} & acımasızlık           & cruauté                   & \cellcolor[HTML]{9AFF99}cruauté       & mé, ence, {[}PAD{]}, {[}PAD{]}             & \cellcolor[HTML]{FFCCC9}méence                    & \cellcolor[HTML]{9AFF99}cruauté        \\
\textbf{DE-IT} & abkürzungen           & abbreviazioni             & \cellcolor[HTML]{9AFF99}abbreviazioni & abbrevi, zioni, {[}PAD{]}, {[}PAD{]}       & \cellcolor[HTML]{FFCCC9}abbrevizioni              & \cellcolor[HTML]{9AFF99}abbreviazioni  \\
\textbf{DE-IT} & antibiotika           & antibiotici               & \cellcolor[HTML]{9AFF99}antibiotici   & antibi, oti, ici, {[}PAD{]}                & \cellcolor[HTML]{FFCCC9}antibiotiici              & \cellcolor[HTML]{9AFF99}antibiotici    \\
\textbf{DE-IT} & bruderschaft          & fratellanza               & \cellcolor[HTML]{FFCCC9}confraternita & confratern, fratern, fratern, .            & \cellcolor[HTML]{FFCCC9}confraternfraternfratern. & \cellcolor[HTML]{FFCCC9}confraternita  \\ \bottomrule
\end{tabular}}
\caption{Translation Examples in the Multilingual Setting. The table displays the language pairs, source words, corresponding target words, and translations predicted by the CLC1, ProMap$_G$, and ProMap$_S$ models, as well as the sub-tokens generated by the ProMap$_G$ model. A green background indicates a correct prediction, while a red background indicates an incorrect prediction.}
\label{tab:predictions-examples-multilingual}
\end{table*}

\begin{table*}[t]
\centering 
\includegraphics[width=0.95\linewidth]{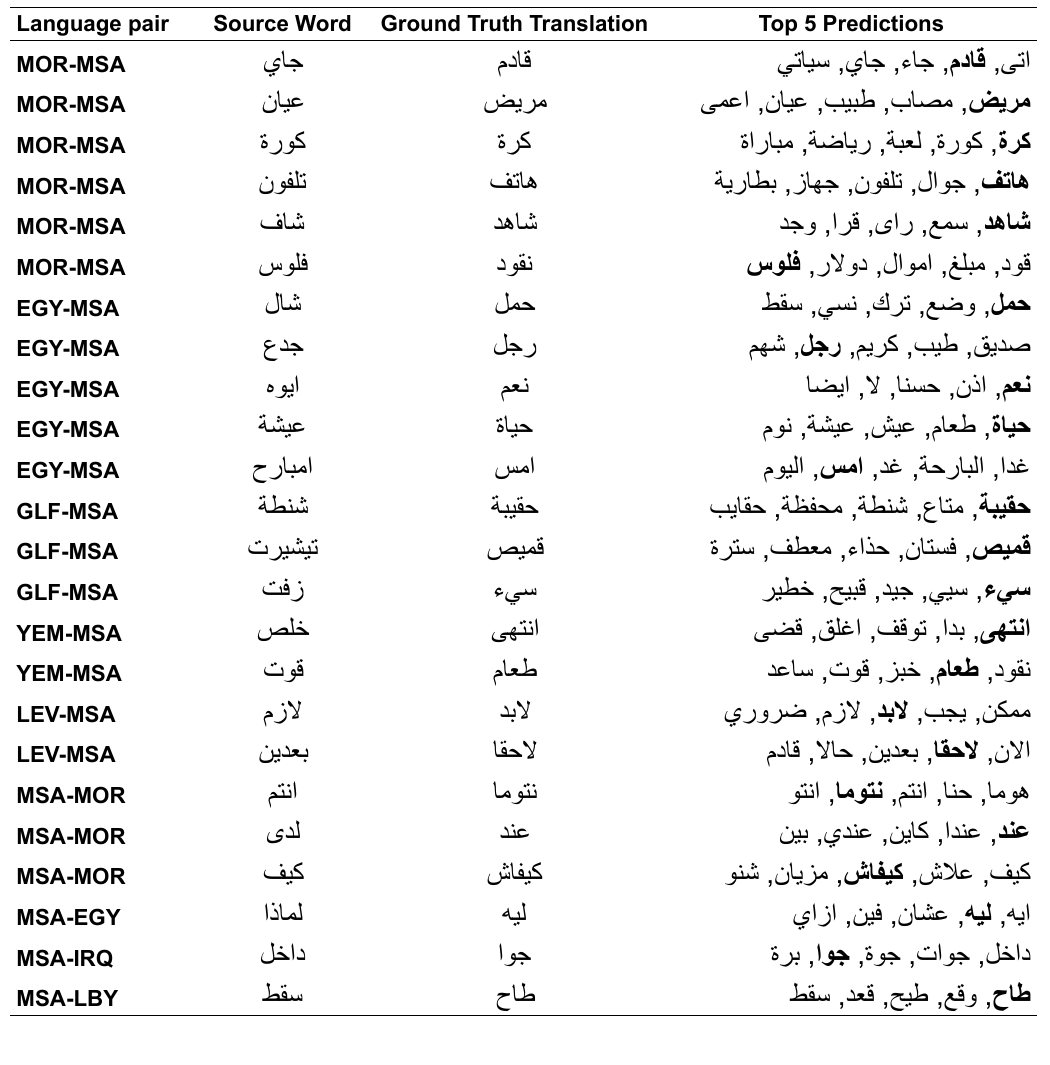}
\caption{Examples of word translations between Arabic dialects and MSA generated by ProMap$_G$ with MARBERT. The table presents several Arabic variant pairs and the top 5 predictions for each query. The top 5 predictions are presented from right to left direction.}
\label{tab:arabic_examples}
\end{table*}

\subsection{Comparison Between Dialectal Arabic PLMs}
\label{sec:plms-compare-app}

We evaluate the performance of ProMap$_G$ for word translation between different Arabic dialects and MSA using two dialectal Arabic PLMs: MARBERT and CAMELBERT (the mix variant). The results, summarized in Table \ref{tab:comparison-between-arabic-plms}, indicate that MARBERT outperformed CAMELBERT in the majority of experiments. Additionally, it is worth noting that CAMELBERT has a vocabulary of 30k tokens, while MARBERT has a vocabulary of 100k tokens. These factors led us to adopt MARBERT for our results in the paper.

It should also be noted that the results presented in Table \ref{tab:shared-dict-results} in the paper differ from those in Table \ref{tab:comparison-between-arabic-plms} because the latter table evaluates the overlapped dictionary pairs between the two PLMs vocabularies, while the results reported in the paper were based on pairs covered by MARBERT vocabulary only.

\begin{table*}[!htb]
\centering
\resizebox{2.\columnwidth}{!}{%
\begin{tabular}{cccccccccccccccccccccc}
\hline
                                  &               & \multicolumn{2}{c}{\textbf{MAR-MSA}} & \multicolumn{2}{c}{\textbf{ALG-MSA}} & \multicolumn{2}{c}{\textbf{TUN-MSA}} & \multicolumn{2}{c}{\textbf{LBY-MSA}}  & \multicolumn{2}{c}{\textbf{EGY-MSA}} & \multicolumn{2}{c}{\textbf{SDN-MSA}}  & \multicolumn{2}{c}{\textbf{LEV-MSA}} & \multicolumn{2}{c}{\textbf{IRQ-MSA}} & \multicolumn{2}{c}{\textbf{GLF-MSA}} & \multicolumn{2}{c}{\textbf{YEM-MSA}} \\
                                  &               & \textbf{$\to$} & \textbf{$\gets$} & \textbf{$\to$} & \textbf{$\gets$} & \textbf{$\to$} & \textbf{$\gets$} & \textbf{$\to$} & \textbf{$\gets$} & \textbf{$\to$} & \textbf{$\gets$} & \textbf{$\to$} & \textbf{$\gets$} & \textbf{$\to$} & \textbf{$\gets$} & \textbf{$\to$} & \textbf{$\gets$} & \textbf{$\to$}  & \textbf{$\gets$} & \textbf{$\to$} & \textbf{$\gets$} \\ \hline
\multirow{4}{*}{\textbf{CAMELBERT}}   & \textbf{P@1}  & 60.71                     & 50.36                  & 65.06                     & 54.08                  & 63.03                     & 56.43                  & 68.99                     & 56.49                  & 62.34                     & 51.40                   & 57.71                     & 44.17                  & 58.33                     & 50.24                  & 73.01                     & 61.58                  & 50.85                      & 32.71                  & 60.21                     & 52.22                  \\
                                  & \textbf{P@5}  & 70.54                     & 61.31                  & 77.11                     & 65.31                  & 73.95                     & 65.00                     & 83.72                     & 75.32                  & 80.52                     & 69.83                  & 82.86                     & 60.19                  & 76.04                     & 65.88                  & 85.28                     & 75.71                  & 70.94                      & 51.88                  & 81.15                     & 68.97                  \\
                                  & \textbf{P@10} & 76.79                     & 65.69                  & 80.72                     & 69.39                  & 78.15                     & 68.57                  & 88.37                     & 79.87                  & 84.42                     & 75.42                  & 85.71                     & 65.53                  & 81.25                     & 73.93                  & 87.73                     & 79.66                  & 79.91                      & 57.14                  & 83.25                     & 74.38                  \\
                                  & \textbf{P@50} & 83.04                     & 73.72                  & 89.16                     & 78.57                  & 85.71                     & 77.86                  & 92.25                     & 85.71                  & 91.56                     & 85.47                  & 91.43                     & 79.61                  & 91.15                     & 81.52                  & 92.64                     & 86.44                  & 88.46                      & 77.44                  & 88.48                     & 85.71                  \\ \hline
\multirow{4}{*}{\textbf{MARBERT}} & \textbf{P@1}  & 59.82                     & 50.36                  & 65.06                     & 53.06                  & 65.55                     & 54.29                  & 72.09                     & 58.44                  & 66.88                     & 51.40                   & 65.14                     & 47.09                  & 62.50                      & 45.97                  & 79.14                     & 51.41                  & 61.54                      & 30.45                  & 63.35                     & 49.75                  \\
                                  & \textbf{P@5}  & 69.64                     & 61.31                  & 78.31                     & 67.35                  & 78.15                     & 66.43                  & 86.82                     & 74.68                  & 81.17                     & 68.16                  & 84.57                     & 61.65                  & 82.29                     & 67.77                  & 87.73                     & 61.02                  & 80.77                      & 49.62                  & 84.29                     & 67.49                  \\
                                  & \textbf{P@10} & 76.79                     & 62.77                  & 79.52                     & 69.39                  & 81.51                     & 70.71                  & 87.60                      & 79.87                  & 85.71                     & 73.74                  & 87.43                     & 67.48                  & 84.90                      & 72.51                  & 91.41                     & 63.28                  & 85.04                      & 58.27                  & 88.48                     & 75.37                  \\
                                  & \textbf{P@50} & 85.71                     & 69.34                  & 87.95                     & 75.51                  & 88.24                     & 78.57                  & 93.02                     & 87.01                  & 92.86                     & 87.15                  & 94.86                     & 85.92                  & 91.67                     & 84.83                  & 93.87                     & 70.62                  & 93.16                      & 74.06                  & 93.19                     & 84.24                  \\ \hline
\end{tabular}}
\caption{A comparison of P@1 for ProMap$_{G}$ using CAMELBERT and MARBERT as Arabic PLMs between Arabic dialects and MSA.}
\label{tab:comparison-between-arabic-plms}
\end{table*}

\subsection{The Effect of the Prompt Template}
\label{sec:effect-prompt-template}

One of the challenges in prompt-based finetuning is constructing the template, particularly in the context of a cross-lingual task. We had to choose whether the template should be in the source language, the target language, a random language, or include special tokens. To address this question, we conduct several experiments where we apply prompt-based finetuning to all language pairs using four different templates: a template written in the source language, a template written in the target language, a template written in English, a template composed of random tokens from various languages, and a template made from special tokens added to the PLM vocabulary. The results presented in table \ref{tab:different-templates} show that the performance gap between the different templates is not significant, but the templates expressed in the source and target languages yielded the best and most stable results.

\begin{table*}[!t]
\centering
\resizebox{2.\columnwidth}{!}{%
\begin{tabular}{cccccc}
\toprule
\textbf{Pairs}               & \textbf{English template} & \textbf{Source language template} & \textbf{Target language template} & \textbf{Random language template} & \textbf{Special Tokens} \\ \midrule
\textbf{DE $\to$ FR} & 58.07 (1.37)              & 59.08 (1.02)                      & \textbf{59.14 (1.22)}             & 57.47 (0.58)                      & 58.98 (1.48)            \\
\textbf{DE $\to$ IT} & 55.67 (0.29)              & \textbf{56.94 (0.85)}             & 56.30 (1.14)                       & 56.83 (0.47)                      & 55.75 (2.38)            \\
\textbf{DE $\to$ RU} & 44.82 (3.47)              & 45.73 (6.05)                      & \textbf{48.90 (1.32)}              & 48.17 (1.49)                      & 48.29 (1.80)             \\
\textbf{DE $\to$ TR} & 51.25 (1.41)              & 53.42 (1.12)                      & \textbf{53.92 (1.83)}             & 52.17 (0.75)                      & 52.92 (2.48)            \\
\textbf{EN $\to$ DE} & -                         & \textbf{74.55 (2.25)}             & 72.06 (2.16)                      & 73.31 (2.25)                      & 73.62 (1.01)            \\
\textbf{EN $\to$ FR} & -                         & 79.83 (0.91)                      & \textbf{80.44 (1.28)}             & 80.17 (0.58)                      & 79.67 (0.89)            \\
\textbf{EN $\to$ IT} & -                         & \textbf{77.61 (1.36)}             & 76.70 (1.13)                       & 72.54 (7.97)                      & 75.41 (0.71)            \\
\textbf{EN $\to$ RU} & -                         & 55.73 (5.78)                      & \textbf{59.16 (2.53)}             & 57.25 (3.70)                       & 56.79 (2.98)            \\
\textbf{EN $\to$ TR} & -                         & \textbf{58.93 (2.65)}             & 57.33 (1.74)                      & 58.00 (0.85)                       & 57.33 (1.92)            \\
\textbf{IT $\to$ FR} & 72.47 (1.78)              & \textbf{73.55 (1.45)}             & 72.54 (2.30)                       & 72.50 (2.48)                       & 72.96 (1.18)            \\
\textbf{RU $\to$ FR} & 49.66 (3.58)              & 51.26 (2.12)                      & 52.18 (1.13)                      & \textbf{52.94 (1.22)}             & 52.44 (1.50)             \\
\textbf{RU $\to$ IT} & 48.20 (3.01)               & 50.36 (1.81)                      & 48.11 (4.30)                       & \textbf{52.25 (1.01)}             & 51.91 (2.28)            \\
\textbf{TR $\to$ FR} & 46.11 (1.25)              & \textbf{49.87 (0.70)}              & 47.32 (1.36)                      & 48.15 (2.98)                      & 48.52 (0.90)             \\
\textbf{TR $\to$ IT} & 46.92 (1.88)              & \textbf{51.44 (1.54)}             & 46.78 (0.79)                      & 49.04 (2.60)                       & 48.08 (1.62)            \\
\textbf{TR $\to$ RU} & 30.31 (2.59)              & \textbf{37.95 (2.30)}              & 34.52 (1.50)                       & 32.88 (2.17)                      & 35.62 (3.40)             \\ \bottomrule
\end{tabular}}
\caption{Comparison of P@1 Scores of ProMap$_G$ using different prompting templates. Every value in the table presents the average (and standard deviation) of 5 runs, corresponding to 5 random seeds.}
\label{tab:different-templates}
\end{table*}

\subsection{Results of ProMap$_{G}$ on the MADAR Lexicon}
\label{sec:madar-full-results}
Table \ref{tab:madar-results} presents the results achieved on the different Arabic variants covered by the MADAR lexicon (11 pairs). P@1, P@5, P@10 and P@50 scores are reported.

\begin{table}[!htb]
\centering
\resizebox{\columnwidth}{!}{%

\begin{tabular}{lcccc}
\toprule
                                \textbf{Pair}   & \textbf{P@1} & \textbf{P@5} & \textbf{P@10} & \textbf{P@50} \\ \cline{2-5} 
\textbf{MAR $\to$ MSA}  & 52.98          & 64.90           & 70.86           & 83.44           \\
\textbf{MSA $\to$ MAR}  & 39.55          & 54.80           & 59.89           & 72.32           \\ \hline
\textbf{ALG $\to$ MSA}  & 50.81          & 68.55          & 75.00              & 79.84           \\
\textbf{MSA $\to$ ALG}  & 37.32          & 51.41          & 63.38           & 71.13           \\ \hline
\textbf{TUN $\to$ MSA}  & 51.55          & 73.29          & 80.12           & 85.09           \\
\textbf{MSA $\to$ TUN}  & 43.62          & 59.57          & 63.30            & 72.34           \\ \hline
\textbf{LBY $\to$ MSA}    & 66.67          & 81.55          & 84.52           & 91.67           \\
\textbf{MSA $\to$ LBY}    & 52.02          & 69.70           & 76.26           & 86.36           \\ \hline
\textbf{EGY $\to$ MSA}     & 61.95          & 80.49          & 84.39           & 92.68           \\
\textbf{MSA $\to$ EGY}     & 41.63          & 67.81          & 74.68           & 85.41           \\ \hline
\textbf{SDN $\to$ MSA}  & 61.32          & 79.01          & 86.01           & 93.00              \\
\textbf{MSA $\to$ SDN}  & 34.57          & 56.13          & 62.08           & 82.53           \\ \hline
\textbf{LEV $\to$ MSA} & 60.37          & 80.65          & 83.41           & 94.93           \\
\textbf{MSA $\to$ LEV} & 40.91          & 64.46          & 73.14           & 87.60            \\ \hline
\textbf{IRQ $\to$ MSA}     & 67.33          & 85.64          & 88.61           & 93.07           \\
\textbf{MSA $\to$ IRQ}     & 49.08          & 70.18          & 75.69           & 84.86           \\ \hline
\textbf{GLF $\to$ MSA}      & 60.17          & 81.10           & 86.05           & 93.02           \\
\textbf{MSA $\to$ GLF}      & 25.75          & 49.32          & 56.44           & 76.71           \\ \hline
\textbf{YEM $\to$ MSA}    & 56.72          & 81.72          & 86.19           & 91.79           \\
\textbf{MSA $\to$ YEM}    & 38.75          & 59.41          & 68.63           & 82.29           \\ \bottomrule
\textbf{Avg.}                           \\ 
\textbf{* $\to$ MSA}                     & 58.99          & 77.69          & 82.52           & 89.85           \\ 
\textbf{MSA $\to$ *}                     & 40.32          & 60.27          & 67.35           & 80.15           \\ \bottomrule
\end{tabular}}
\caption{Results of ProMap$_G$ for word translation between Arabic dialects and MSA using MARBERT on the MADAR Lexicon.}
\label{tab:madar-results}
\end{table}

\subsection{ProMap$_{G}$ Few-shot Results}
\label{sec:few-shot-res}

\begin{table}[!htb]
\centering
\resizebox{\columnwidth}{!}{%
\begin{tabular}{lll}
\toprule
\textbf{Pair}                   & \textbf{Training example}  & \textbf{P@1} \\ \midrule
\multirow{2}{*}{\textbf{DE-FR}} & zahlung  - paiement        & 15.05           \\
                                & system - système           & 2.15            \\ \hline
\multirow{2}{*}{\textbf{DE-IT}} & expedition - spedizione    & 2.55            \\
                                & fenster - finestra         & 4.88            \\ \hline
\multirow{3}{*}{\textbf{EN-FR}} & jurisdiction - juridiction & 52.35           \\
                                & ideal - idéal              & 8.59            \\
                                & orientation - orientation  & 0.83            \\ \hline
\multirow{2}{*}{\textbf{EN-IT}} & weight - peso              & 15.90            \\
                                & rice - riso                & 1.53            \\ \hline
\multirow{3}{*}{\textbf{EN-TR}} & league - lig               & 31.90            \\
                                & agreement - anlaşma        & 3.81            \\
                                & influence - etki           & 0.00               \\ \hline
\multirow{2}{*}{\textbf{TR-IT}} & olasılık - possibilità     & 6.51            \\
                                & mineral - minerale         & 0.00               \\ \bottomrule
\end{tabular}}
\caption{The effect on the selected training example for the few-shot scenario when $N=1$. The table shows examples of the selected training example for several pairs and the corresponding P@1 score of ProMap$_G$ on the test set.}
\label{tab:one-shot-training-examples}
\end{table}

Tables \ref{tab:few-shot-full-results} and \ref{tab:few-shot-full-results-top-5} show the results of few-shot experiments on ProMap$_G$ for 15 different language pairs. The scores reported are the average of 25 runs with 5 different random samplings of $N$ examples and 5 random seeds. The standard deviation is also reported and it is observed that it is large for many experiments. This is likely due to the choice of training samples for the ProMap$_G$ model. To further investigate this, Table \ref{tab:one-shot-training-examples} presents examples of training samples for the one-shot scenario (where $N=1$) and shows how the chosen sample can greatly impact the performance on the test set. For example, when using the sample \textit{"jurisdiction" - "juridiction"} for the English-French language pair, the P@1 score is 52.35, while the sample \textit{"ideal" - "idéal"} results in a P@1 score of 8.59. In some cases, the chosen training sample can prevent the model from converging at all. This can be seen in the Turkish-Italian pair, where the sample \textit{"mineral" - "minerale"} results in a P@1 score of 0.00, while the sample \textit{"olasılık" - "possibilità"} results in a P@1 score of 6.51. It is also observed that when the chosen example is exclusive to the language pair, the performance is better than when the example is shared with other language pairs.

\begin{table*}[]
\centering
\resizebox{2.\columnwidth}{!}{%
\begin{tabular}{llllllllllllllll}
\toprule
\textbf{Pairs} & \textbf{DE $\to$ FR} & \textbf{DE $\to$ IT} & \textbf{DE $\to$ RU} & \textbf{DE $\to$ TR} & \textbf{EN $\to$ DE} & \textbf{EN $\to$ FR} & \textbf{EN $\to$ IT} & \textbf{EN $\to$ RU} & \textbf{EN $\to$ TR} & \textbf{IT $\to$ FR} & \textbf{RU $\to$ FR} & \textbf{RU $\to$ IT} & \textbf{TR $\to$ FR} & \textbf{TR $\to$ IT} & \textbf{TR $\to$ RU} \\ \midrule
\textbf{N=1}   & 5.42 (1.92)                  & 5.34 (2.86)                  & 0.00 (0.00)                  & 4.21 (3.90)                   & 10.95 (5.46)                 & 20.67 (15.58)                & 11.04 (2.34)                 & 6.87 (1.53)                  & 8.09 (7.05)                  & 13.05 (10.88)                & 6.27 (3.19)                  & 9.70 (7.99)                   & 2.60 (0.59)                   & 7.02 (6.72)                  & 2.74 (0.00)                   \\
\textbf{N=3}   & 22.03 (8.59)                 & 14.99 (6.27)                 & 6.63 (6.82)                  & 7.45 (3.87)                  & 24.50 (8.00)                   & 39.89 (8.92)                 & 32.39 (17.51)                & 12.15 (5.51)                 & 16.80 (9.14)                  & 38.87 (7.81)                 & 17.61 (7.61)                 & 22.4 (1.93)                  & 15.40 (2.46)                  & 10.44 (3.09)                 & 4.43 (2.68)                  \\
\textbf{N=5}   & 23.27 (12.13)                & 29.40 (4.33)                  & 13.42 (5.42)                 & 16.42 (8.44)                 & 39.52 (7.66)                 & 48.08 (16.50)                 & 44.54 (8.78)                 & 31.08 (2.87)                 & 22.48 (6.71)                 & 48.24 (8.64)                 & 25.81 (8.34)                 & 25.29 (2.37)                 & 23.03 (6.54)                 & 17.73 (7.74)                 & 7.64 (3.75)                  \\
\textbf{N=10}  & 31.10 (6.88)                  & 25.27 (13.35)                & 19.88 (5.03)                 & 26.68 (10.79)                & 52.71 (2.93)                 & 66.03 (6.50)                  & 52.85 (7.04)                 & 32.04 (4.15)                 & 28.55 (11.40)                 & 52.25 (4.36)                 & 31.98 (5.46)                 & 30.93 (0.94)                 & 33.44 (1.32)                 & 28.95 (4.71)                 & 16.87 (3.26)                 \\
\textbf{N=16}  & 30.46 (9.54)                 & 18.74 (6.69)                 & 22.99 (3.32)                 & 29.43 (7.81)                 & 45.44 (6.94)                 & 65.40 (5.96)                  & 51.17 (7.89)                 & 35.60 (5.01)                  & 34.60 (7.35)                  & 39.57 (11.49)                & 35.26 (2.41)                 & 33.10 (1.85)                  & 36.50 (0.86)                  & 34.75 (1.51)                 & 25.84 (2.15)                 \\
\textbf{N=32}  & 37.72 (8.9)                  & 34.01 (1.37)                 & 27.47 (4.94)                 & 26.79 (3.44)                 & 53.63 (4.08)                 & 59.45 (9.21)                 & 47.17 (17.87)                & 42.26 (4.96)                 & 37.68 (5.81)                 & 46.11 (6.37)                 & 40.35 (2.22)                 & 36.29 (0.61)                 & 39.06 (1.64)                 & 39.47 (2.20)                  & 28.16 (1.56)                 \\
\textbf{N=64}  & 36.61 (4.14)                 & 38.34 (4.35)                 & 28.16 (4.86)                 & 30.57 (9.11)                 & 55.45 (6.49)                 & 59.8 (10.21)                 & 52.83 (13.38)                & 46.49 (2.50)                  & 41.40 (11.4)                  & 55.57 (6.67)                 & 41.97 (2.61)                 & 38.40 (2.42)                  & 41.29 (1.16)                 & 43.55 (0.64)                 & 30.66 (1.73)                 \\
\textbf{N=128} & 46.85 (3.44)                 & 45.95 (2.13)                 & 27.04 (6.06)                 & 32.77 (8.96)                 & 54.84 (13.3)                 & 70.44 (3.48)                 & 56.22 (11.46)                & 31.88 (17.06)                & 42.85 (4.34)                 & 59.06 (9.14)                 & 39.12 (13.12)                & 42.81 (2.25)                 & 40.96 (0.72)                 & 44.64 (0.65)                 & 32.98 (1.50)                  \\
\textbf{N=256} & 53.18 (2.63)                 & 50.82 (2.44)                 & 37.38 (6.02)                 & 46.64 (0.95)                 & 61.69 (6.22)                 & 74.37 (1.98)                 & 66.33 (3.35)                 & 45.44 (7.99)                 & 48.94 (2.62)                 & 68.74 (2.25)                 & 49.78 (1.77)                 & 45.37 (2.11)                 & 44.59 (0.91)                 & 47.00 (1.52)                  & 33.75 (1.87)                 \\
\textbf{N=512} & 53.69 (4.02)                 & 51.33 (6.86)                 & 42.14 (4.22)                 & 48.46 (5.74)                 & 67.44 (4.21)                 & 74.21 (4.48)                 & 65.96 (10.78)                & 51.55 (4.65)                 & 57.17 (7.06)                 & 66.11 (8.81)                 & 51.57 (3.35)                 & 47.2 (2.21)                  & 46.24 (0.65)                 & 44.75 (6.61)                 & 36.20 (0.70)                   \\ \bottomrule
\end{tabular}

}
\caption{Comparison of P@1 scores of ProMap$_G$ using different sizes of training example pairs. Every value in the table presents the average (and standard deviation) of 25 runs, corresponding to 5 random samplings with 5 random seeds.}
\label{tab:few-shot-full-results}
\end{table*}

\begin{table*}[]
\centering
\resizebox{2.\columnwidth}{!}{%
\begin{tabular}{llllllllllllllll}
\toprule
\textbf{Pairs} & \textbf{DE $\to$ FR} & \textbf{DE $\to$ IT} & \textbf{DE $\to$ RU} & \textbf{DE $\to$ TR} & \textbf{EN $\to$ DE} & \textbf{EN $\to$ FR} & \textbf{EN $\to$ IT} & \textbf{EN $\to$ RU} & \textbf{EN $\to$ TR} & \textbf{IT $\to$ FR} & \textbf{RU $\to$ FR} & \textbf{RU $\to$ IT} & \textbf{TR $\to$ FR} & \textbf{TR $\to$ IT} & \textbf{TR $\to$ RU} \\ \midrule
\textbf{N=1}   & 10.97 (4.75)                 & 10.56 (3.71)                 & 4.41 (2.71)                  & 11.02 (8.03)                 & 21.30 (7.82)                  & 30.77 (13.02)                & 24.25 (4.34)                 & 9.21 (5.77)                  & 15.22 (5.65)                 & 30.48 (9.70)                  & 23.01 (9.65)                 & 24.36 (12.63)                & 9.22 (6.23)                  & 11.50 (3.65)                  & 7.27 (1.89)                  \\
\textbf{N=3}   & 39.30 (9.39)                  & 26.14 (12.01)                & 22.74 (7.01)                 & 12.85 (7.01)                 & 33.74 (13.18)                & 46.03 (10.55)                & 46.46 (25.67)                & 24.12 (12.09)                & 30.43 (13.26)                & 39.84 (9.89)                 & 37.28 (16.67)                & 44.51 (3.38)                 & 29.19 (8.14)                 & 22.84 (6.10)                  & 9.81 (2.64)                  \\
\textbf{N=5}   & 36.53 (19.20)                 & 30.27 (16.60)                 & 30.34 (11.29)                & 24.11 (8.97)                 & 49.02 (16.12)                & 60.32 (23.94)                & 53.44 (8.22)                 & 35.29 (19.58)                & 38.14 (13.32)                & 70.09 (19.96)                & 49.61 (11.49)                & 48.48 (4.45)                 & 40.98 (8.18)                 & 35.07 (9.62)                 & 16.50 (4.17)                  \\
\textbf{N=10}  & 37.58 (12.83)                & 28.15 (18.80)                 & 34.84 (8.76)                 & 35.76 (16.70)                 & 75.91 (2.72)                 & 82.63 (8.08)                 & 70.59 (8.49)                 & 53.70 (6.96)                  & 41.59 (18.41)                & 61.19 (10.24)                & 57.56 (6.76)                 & 54.36 (4.27)                 & 52.63 (2.92)                 & 49.04 (5.55)                 & 31.84 (4.54)                 \\
\textbf{N=16}  & 44.60 (19.01)                 & 28.35 (11.78)                & 38.73 (9.73)                 & 35.12 (13.23)                & 63.83 (9.13)                 & 69.52 (14.17)                & 67.19 (9.46)                 & 56.16 (7.17)                 & 53.69 (13.17)                & 49.89 (20.10)                 & 62.56 (3.56)                 & 56.02 (5.42)                 & 57.19 (0.77)                 & 54.98 (1.93)                 & 42.33 (2.22)                 \\
\textbf{N=32}  & 49.55 (12.27)                & 48.06 (4.82)                 & 37.04 (12.4)                 & 38.29 (10.7)                 & 62.67 (15.82)                & 73.97 (11.43)                & 56.77 (21.09)                & 60.35 (9.55)                 & 58.03 (10.85)                & 56.02 (15.95)                & 67.36 (1.23)                 & 61.13 (1.39)                 & 60.21 (2.39)                 & 60.20 (2.55)                  & 47.26 (2.89)                 \\
\textbf{N=64}  & 56.86 (6.43)                 & 60.86 (8.09)                 & 45.45 (6.48)                 & 42.28 (14.55)                & 73.80 (8.67)                  & 73.59 (11.51)                & 68.17 (19.03)                & 57.32 (17.71)                & 54.78 (12.92)                & 70.02 (12.97)                & 70.74 (0.51)                 & 66.31 (1.34)                 & 63.62 (1.17)                 & 66.00 (1.79)                  & 53.50 (0.72)                  \\
\textbf{N=128} & 69.69 (4.81)                 & 70.54 (4.01)                 & 46.97 (11.62)                & 49.50 (14.17)                 & 69.60 (18.88)                 & 78.62 (15.37)                & 76.51 (11.56)                & 49.67 (21.26)                & 50.55 (16.40)                 & 71.24 (21.90)                 & 64.23 (17.46)                & 61.98 (13.20)                 & 51.30 (24.37)                 & 68.24 (0.47)                 & 55.84 (0.72)                 \\
\textbf{N=256} & 76.40 (2.12)                  & 75.98 (1.60)                  & 56.96 (11.17)                & 69.11 (1.86)                 & 83.75 (6.13)                 & 88.94 (2.00)                  & 85.95 (2.98)                 & 68.21 (5.48)                 & 67.94 (6.27)                 & 85.99 (0.84)                 & 73.33 (0.51)                 & 70.52 (1.56)                 & 66.67 (1.63)                 & 71.34 (1.3)                  & 56.41 (1.15)                 \\
\textbf{N=512} & 76.36 (3.13)                 & 76.45 (6.15)                 & 66.78 (3.27)                 & 69.68 (6.46)                 & 87.12 (2.37)                 & 89.70 (1.71)                  & 85.19 (9.52)                 & 71.62 (3.87)                 & 77.81 (4.63)                 & 83.52 (5.19)                 & 73.8 (1.90)                   & 71.60 (1.55)                  & 70.21 (0.76)                 & 69.41 (7.06)                 & 59.66 (0.45)                 \\ \bottomrule
\end{tabular}}
\caption{Comparison of P@5 scores of ProMap$_G$ using different sizes of training example pairs. Every value in the table presents the average (and standard deviation) of 25 runs, corresponding to 5 random samplings with 5 random seeds.}
\label{tab:few-shot-full-results-top-5}
\end{table*}

\subsection{Visualisations of the Word Embeddings generated by the mPLM Before and After Finetuning}
\label{sec:visu-tsne}

\begin{figure*}[p!]
\centering
\noindent\begin{minipage}[b]{0.45\textwidth}
\centering
\includegraphics[scale=0.2]{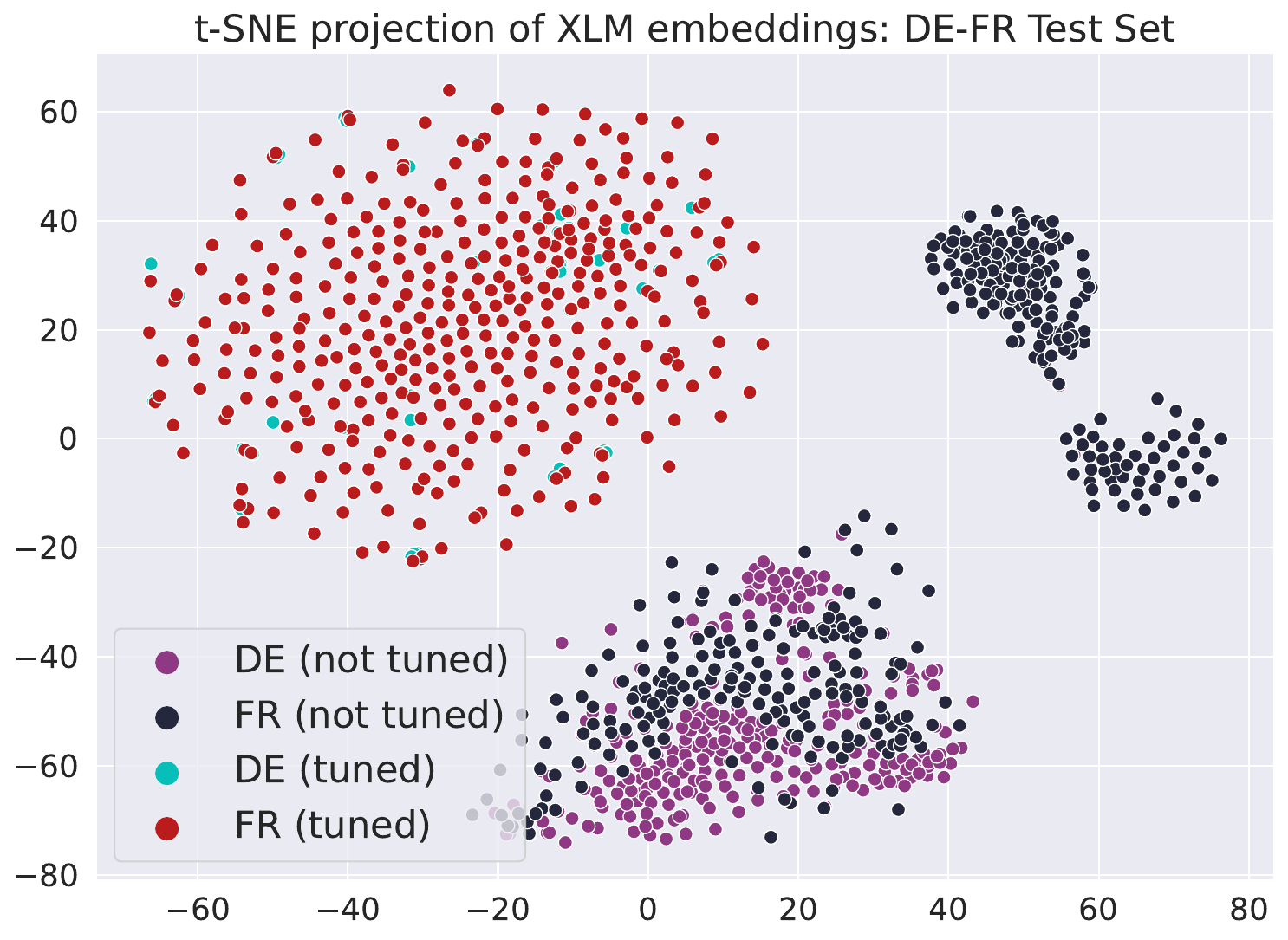}
\caption{A t-SNE visualization of word embeddings generated from the mPLM for words in the DE-FR pair test set, before and after the prompt-based finetuning.}
\label{fig:tsne-appendix-1}
\end{minipage}%
\hfill\begin{minipage}[b]{0.45\textwidth}
\centering
    \includegraphics[scale=0.2]{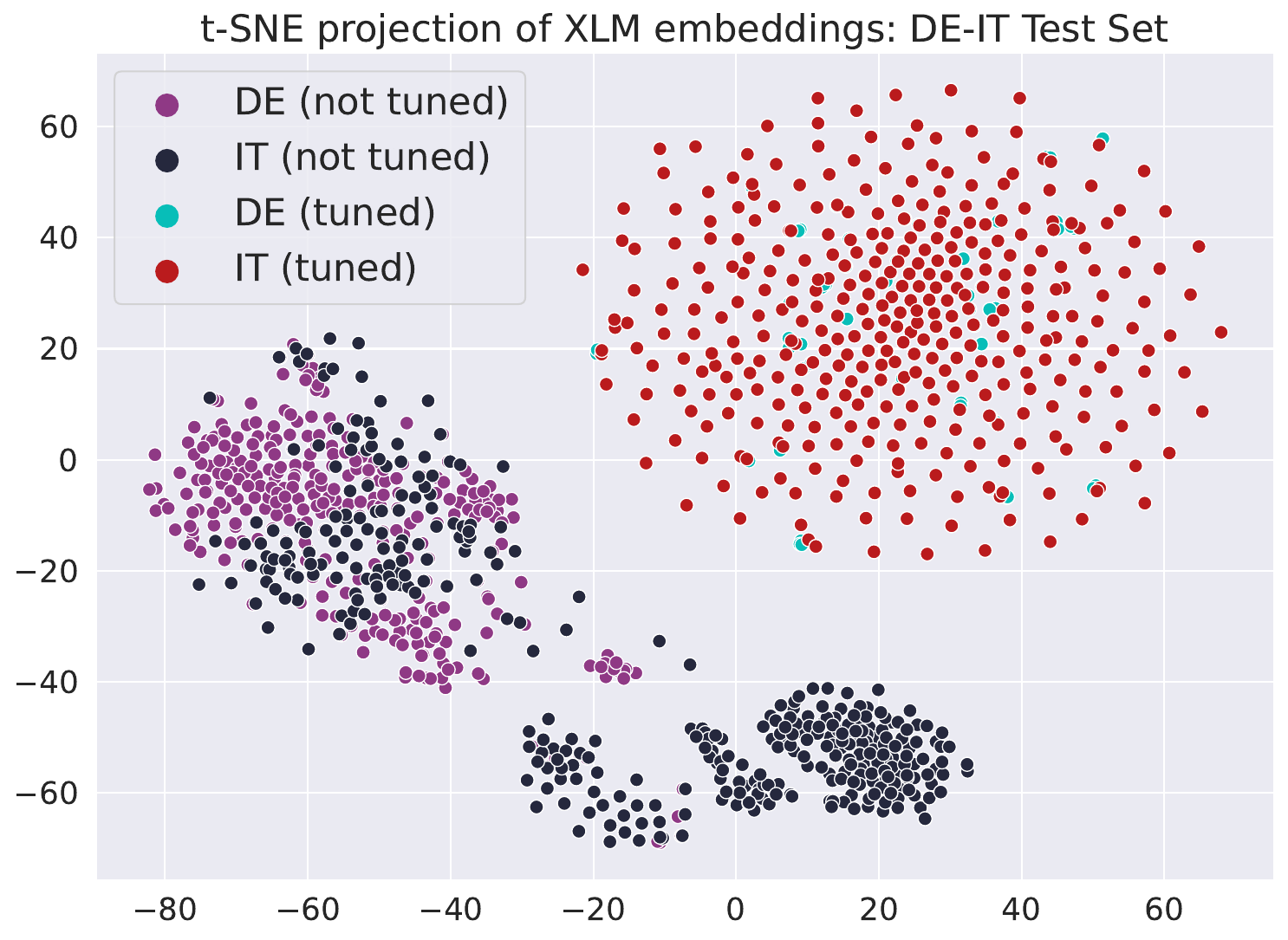}
    \caption{A t-SNE visualization of word embeddings generated from the mPLM for words in the DE-IT pair test set, before and after the prompt-based finetuning.}
\end{minipage}%

\end{figure*}

\begin{figure*}[ht!]
\centering
\noindent\begin{minipage}[b]{0.3\textwidth}
\centering
\includegraphics[scale=0.2]{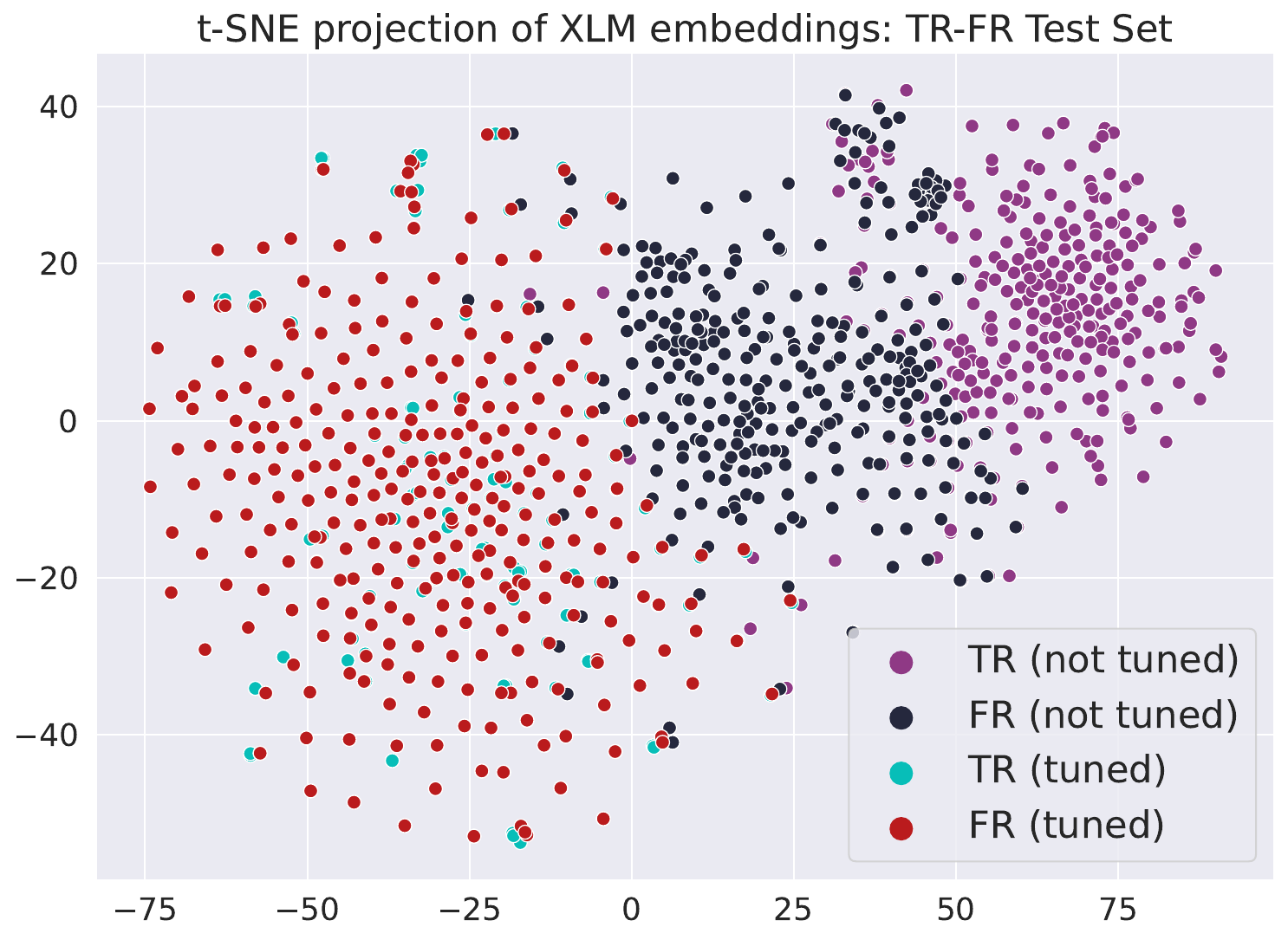}
\caption{A t-SNE visualization of word embeddings generated from the mPLM for words in the TR-FR pair test set, before and after the prompt-based finetuning.}
\end{minipage}%
\hfill\begin{minipage}[b]{0.3\textwidth}
\centering
    \includegraphics[scale=0.2]{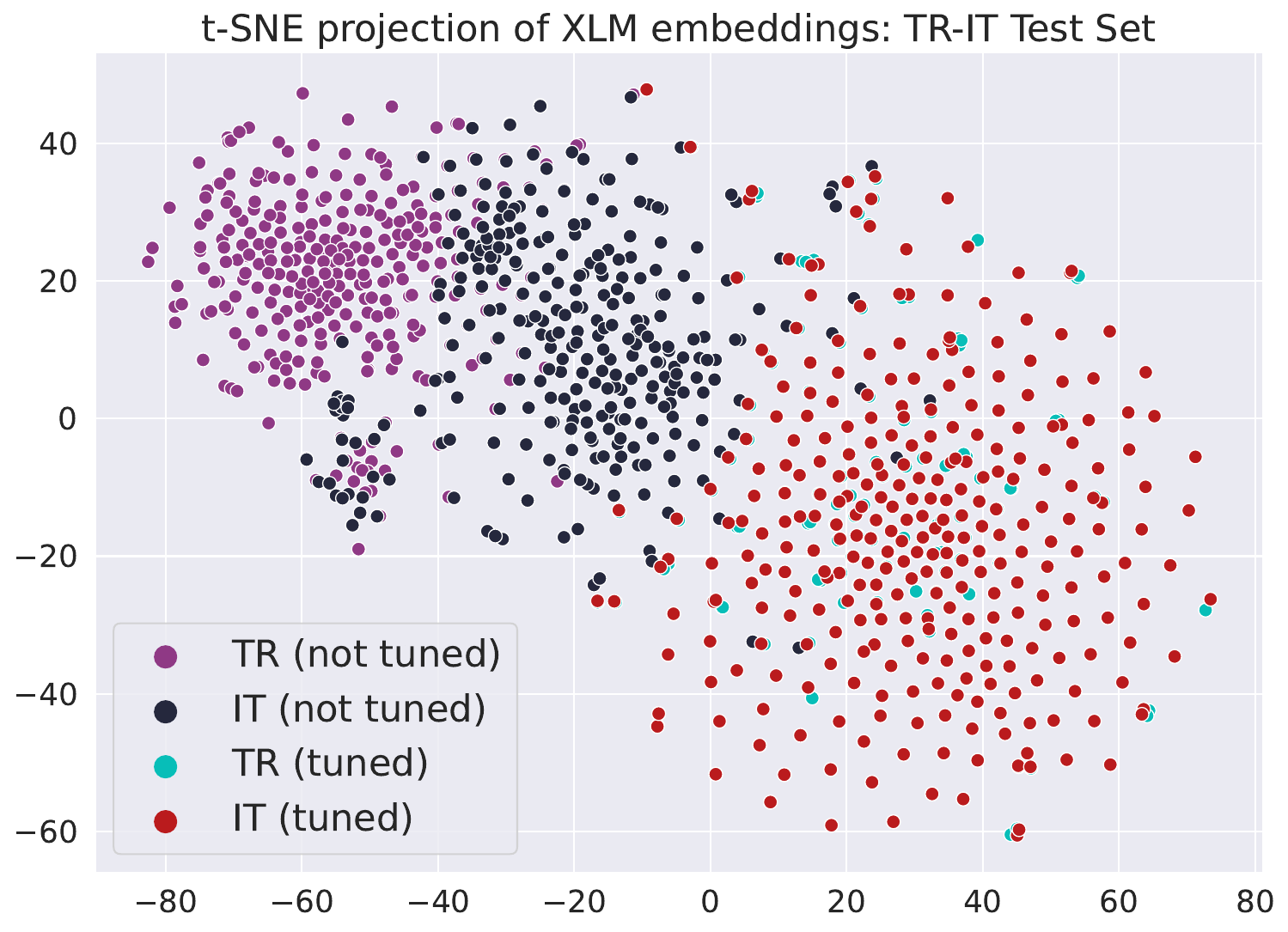}
    \caption{A t-SNE visualization of word embeddings generated from the mPLM for words in the TR-IT pair test set, before and after the prompt-based finetuning.}
\end{minipage}%
\hfill\begin{minipage}[b]{0.3\textwidth}
\centering
    \includegraphics[scale=0.2]{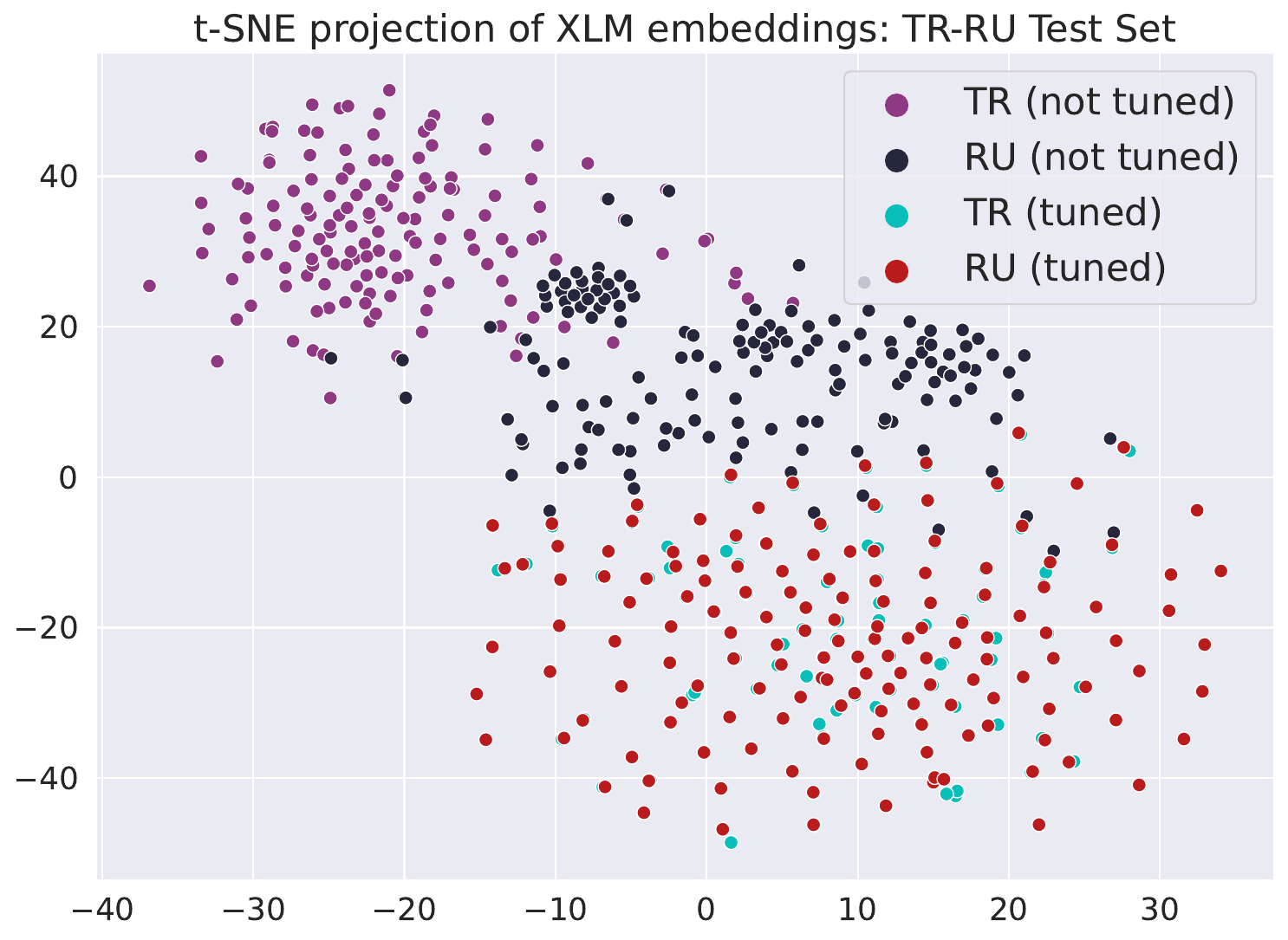}
    \caption{A t-SNE visualization of word embeddings generated from the mPLM for words in the TR-RU pair test set, before and after the prompt-based finetuning.}
    \label{fig:tsne-appendix-last}

\end{minipage}%
\end{figure*}

Figures (\ref{fig:tsne-appendix-1}-\ref{fig:tsne-appendix-last}) present the t-SNE visualizations of the XLM-17 embeddings generated for the word pairs available in the test sets for the different language pairs before and after the prompt-based finetuning using ProMap$_G$. 

\begin{figure*}[ht!]
\centering
\noindent\begin{minipage}[b]{0.45\textwidth}
\centering
\includegraphics[scale=0.2]{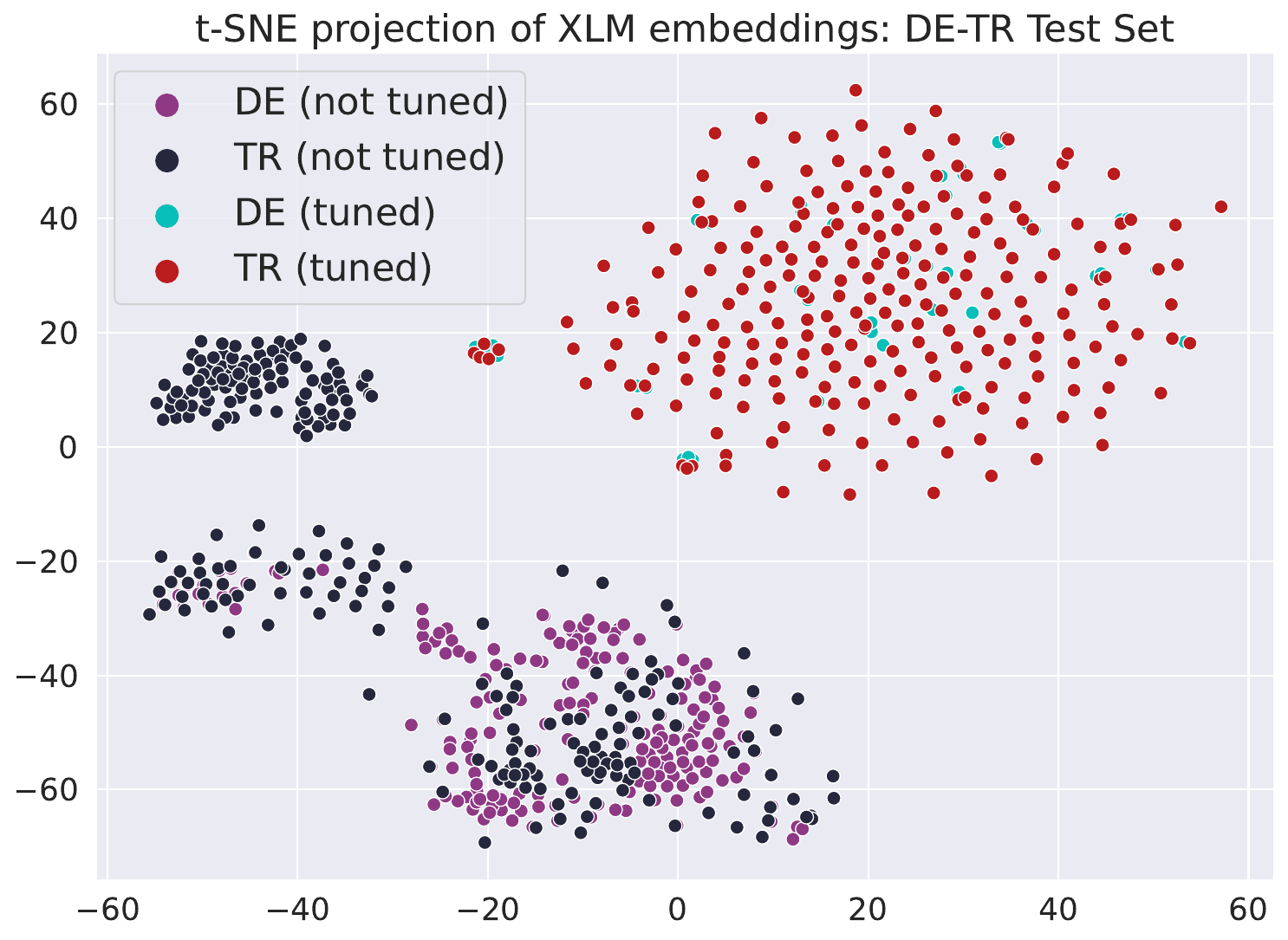}
\caption{A t-SNE visualization of word embeddings generated from the mPLM for words in the DE-TR pair test set, before and after the prompt-based finetuning.}
\end{minipage}%
\hfill\begin{minipage}[b]{0.45\textwidth}
\centering
    \includegraphics[scale=0.2]{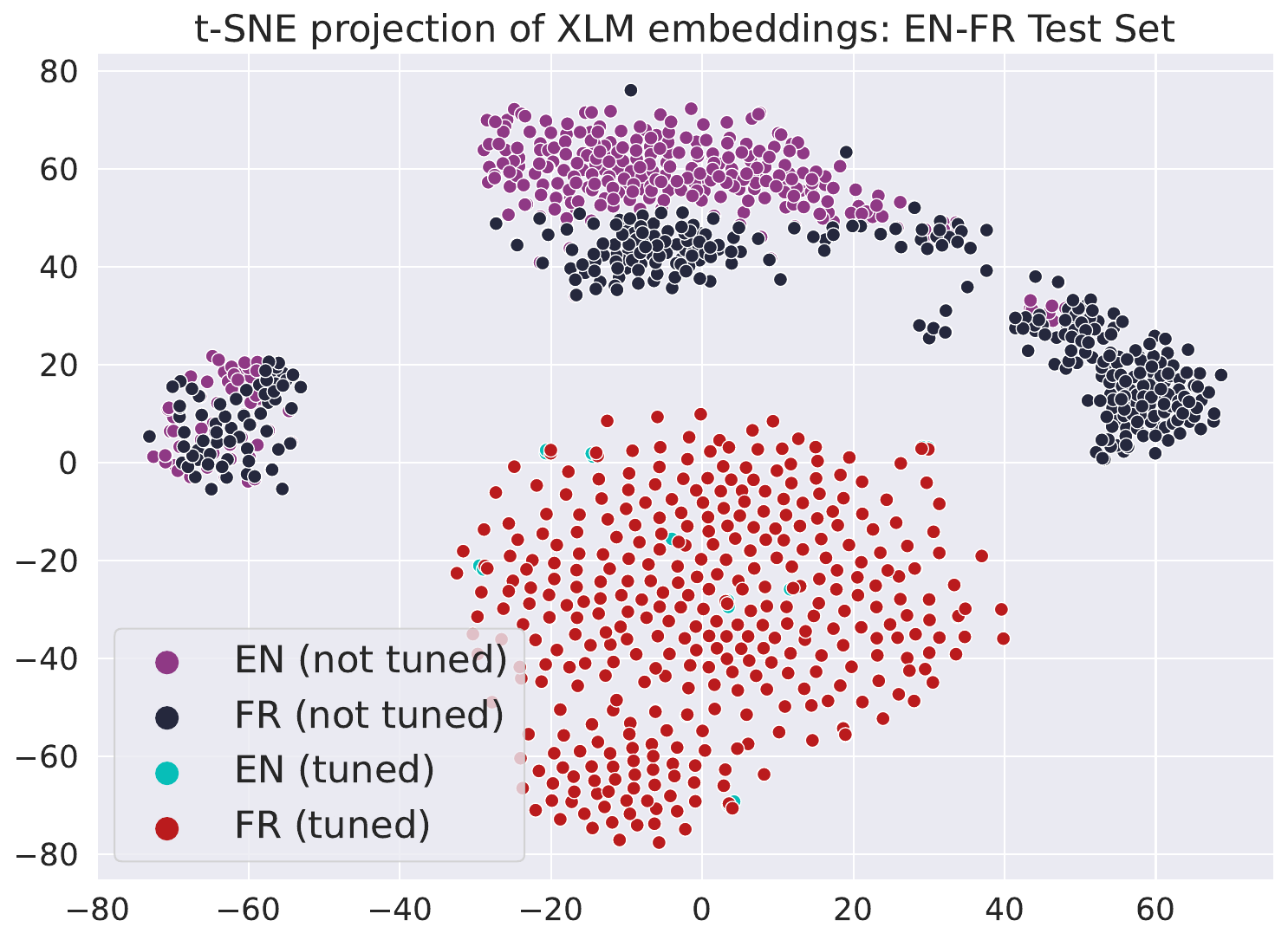}
    \caption{A t-SNE visualization of word embeddings generated from the mPLM for words in the EN-FR pair test set, before and after the prompt-based finetuning.}
\end{minipage}%
\end{figure*}

\begin{figure*}[ht!]
\centering
\noindent\begin{minipage}[b]{0.3\textwidth}
\centering
\includegraphics[scale=0.2]{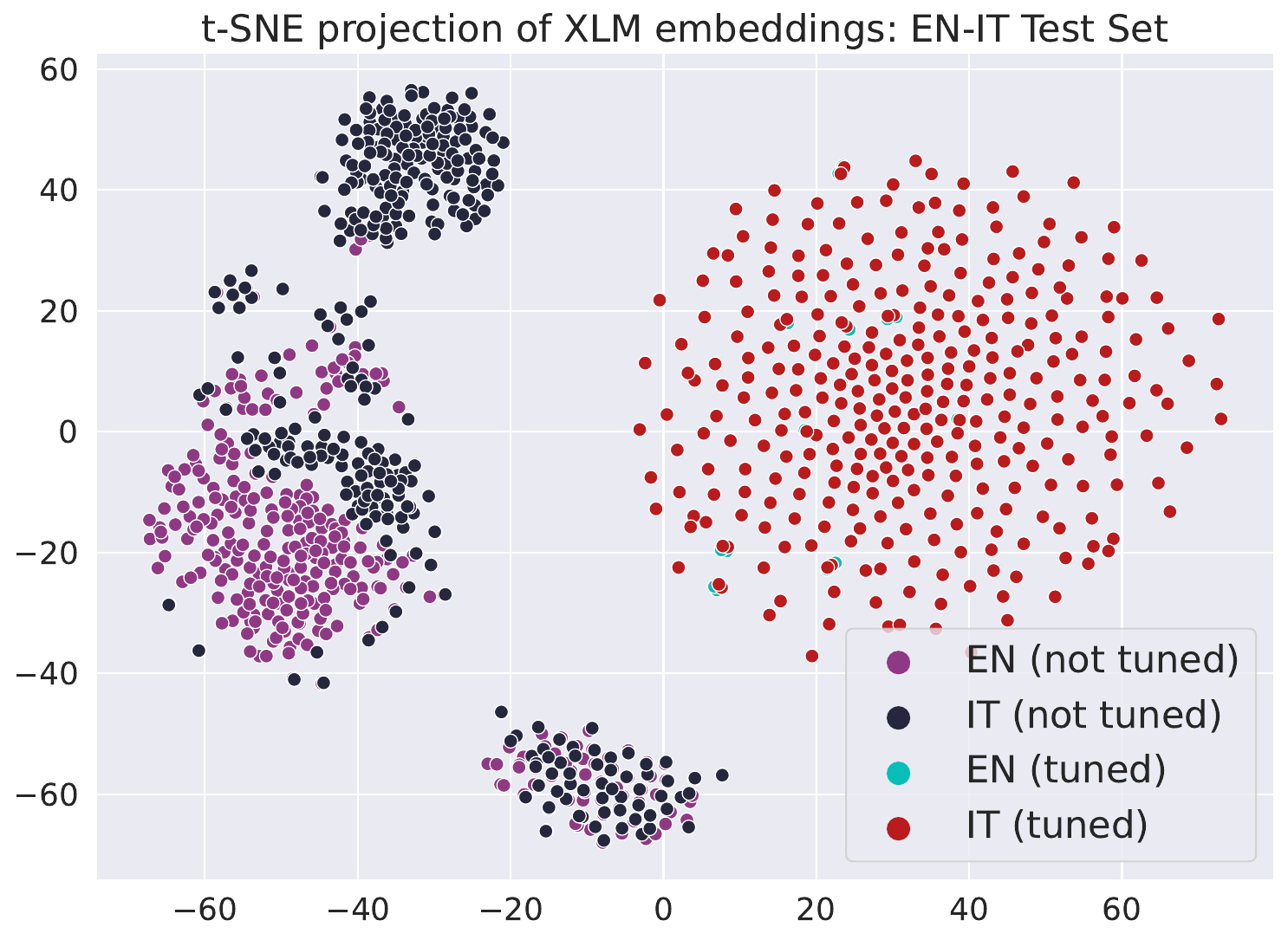}
\caption{A t-SNE visualization of word embeddings generated from the mPLM for words in the EN-IT pair test set, before and after the prompt-based finetuning.}
\end{minipage}%
\hfill\begin{minipage}[b]{0.3\textwidth}
\centering
    \includegraphics[scale=0.2]{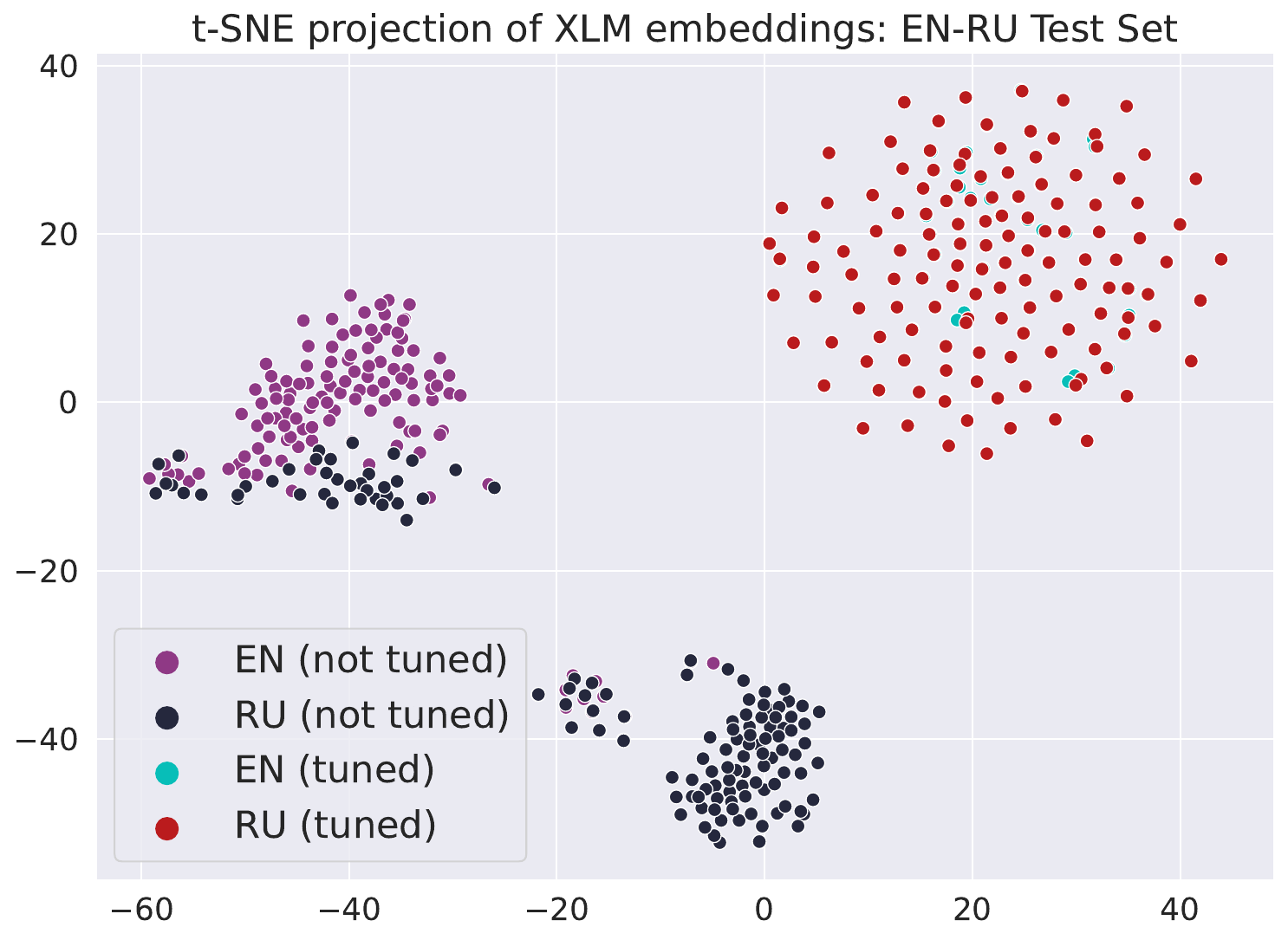}
    \caption{A t-SNE visualization of word embeddings generated from the mPLM for words in the EN-RU pair test set, before and after the prompt-based finetuning.}
\end{minipage}%
\hfill\begin{minipage}[b]{0.3\textwidth}
\centering
    \includegraphics[scale=0.2]{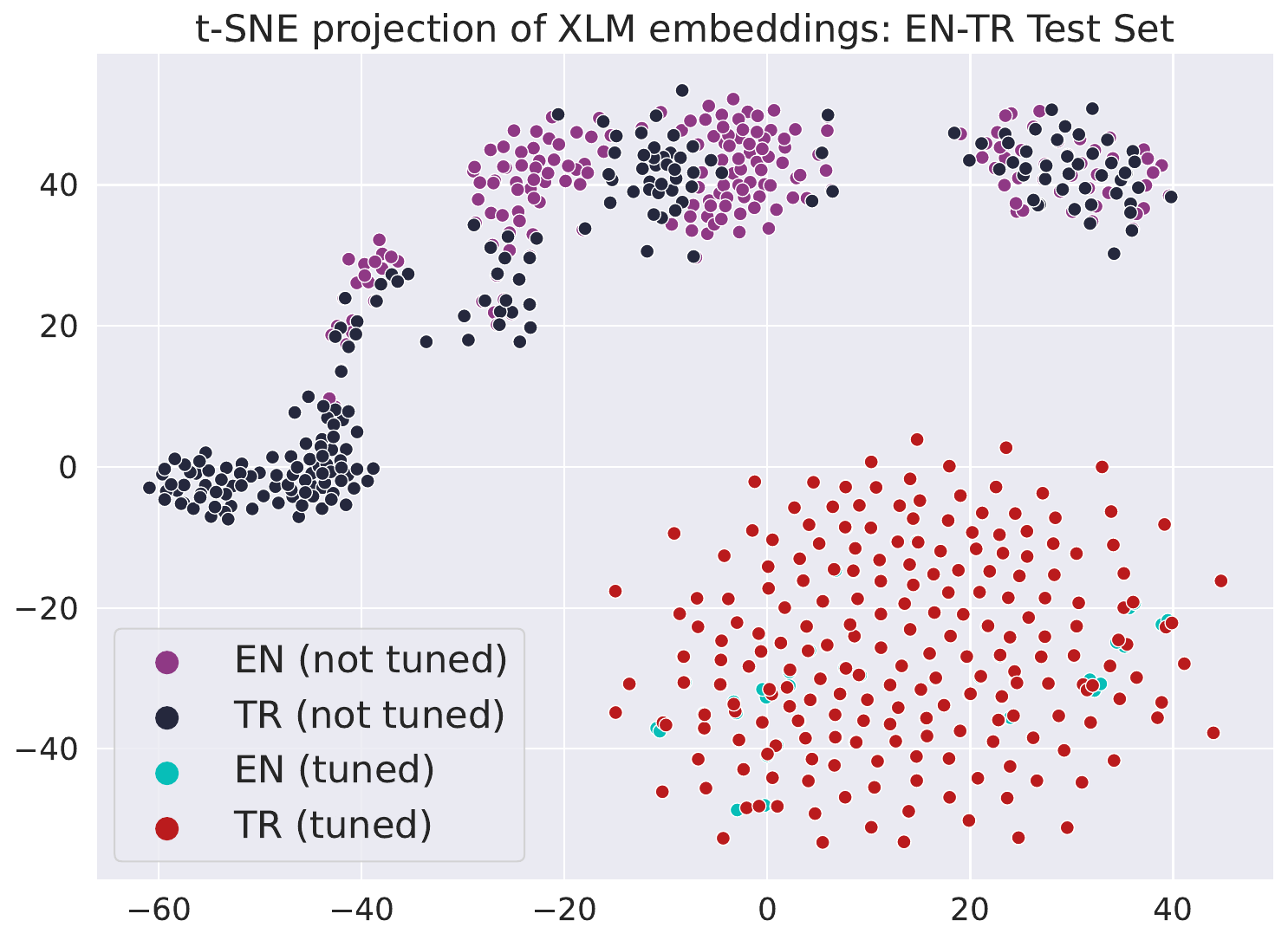}
    \caption{A t-SNE visualization of word embeddings generated from the mPLM for words in the EN-TR pair test set, before and after the prompt-based finetuning.}
\end{minipage}%
\end{figure*}

\begin{figure*}[ht!]
\centering
\noindent\begin{minipage}[b]{0.3\textwidth}
\centering
\includegraphics[scale=0.2]{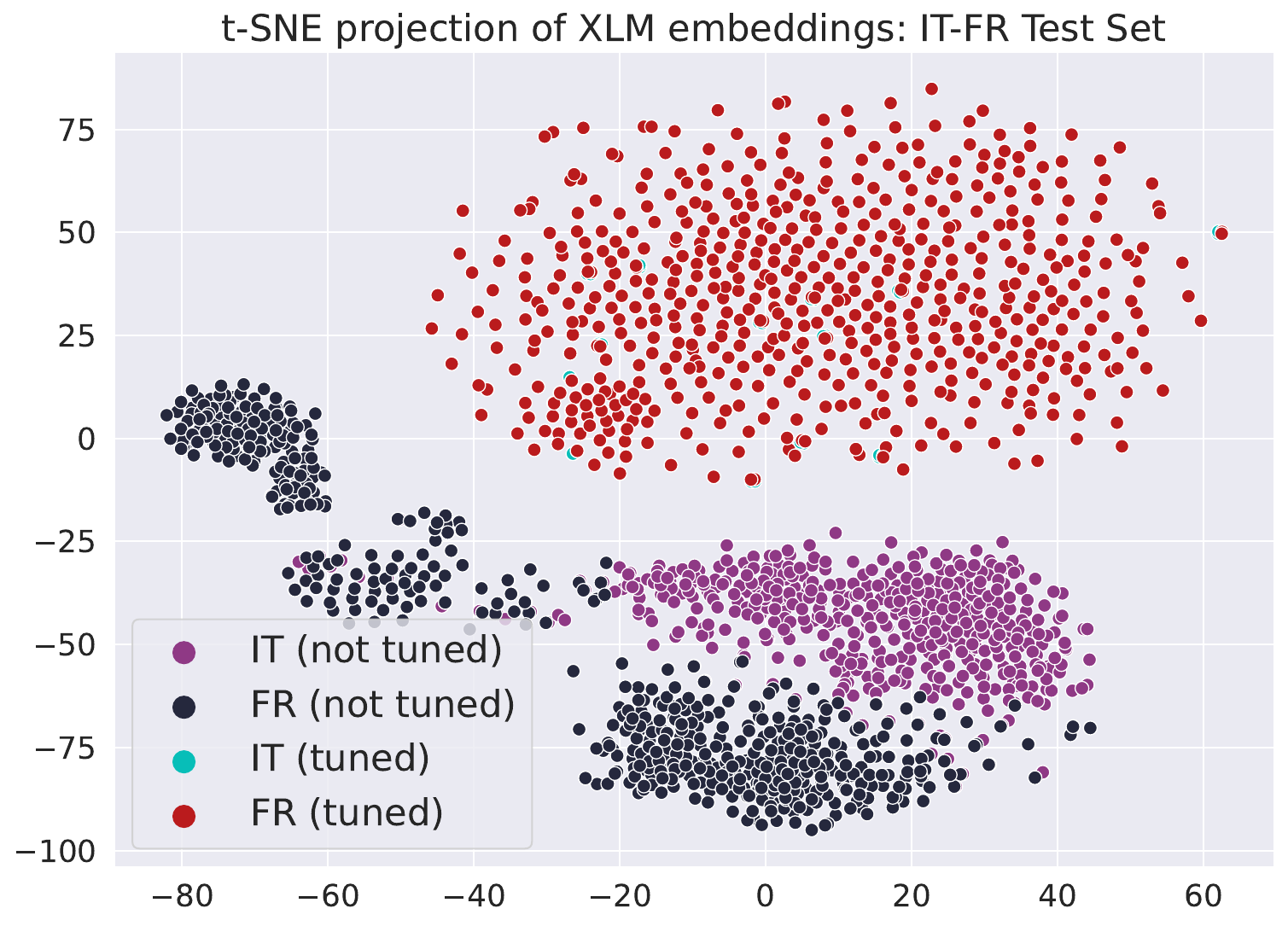}
\caption{A t-SNE visualization of word embeddings generated from the mPLM for words in the IT-FR pair test set, before and after the prompt-based finetuning.}
\end{minipage}%
\hfill\begin{minipage}[b]{0.3\textwidth}
\centering
    \includegraphics[scale=0.2]{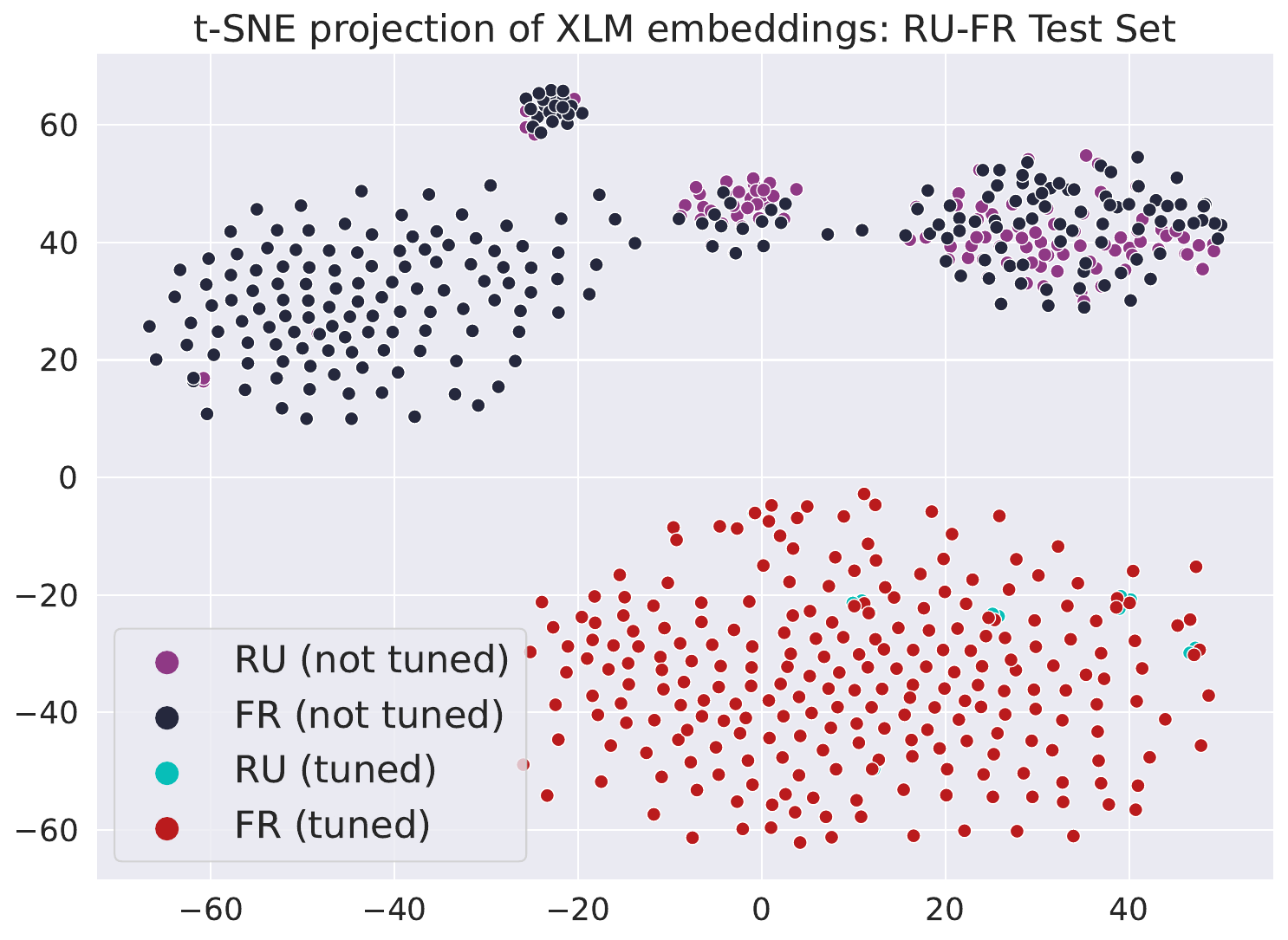}
    \caption{A t-SNE visualization of word embeddings generated from the mPLM for words in the RU-FR pair test set, before and after the prompt-based finetuning.}
\end{minipage}%
\hfill\begin{minipage}[b]{0.3\textwidth}
\centering
    \includegraphics[scale=0.2]{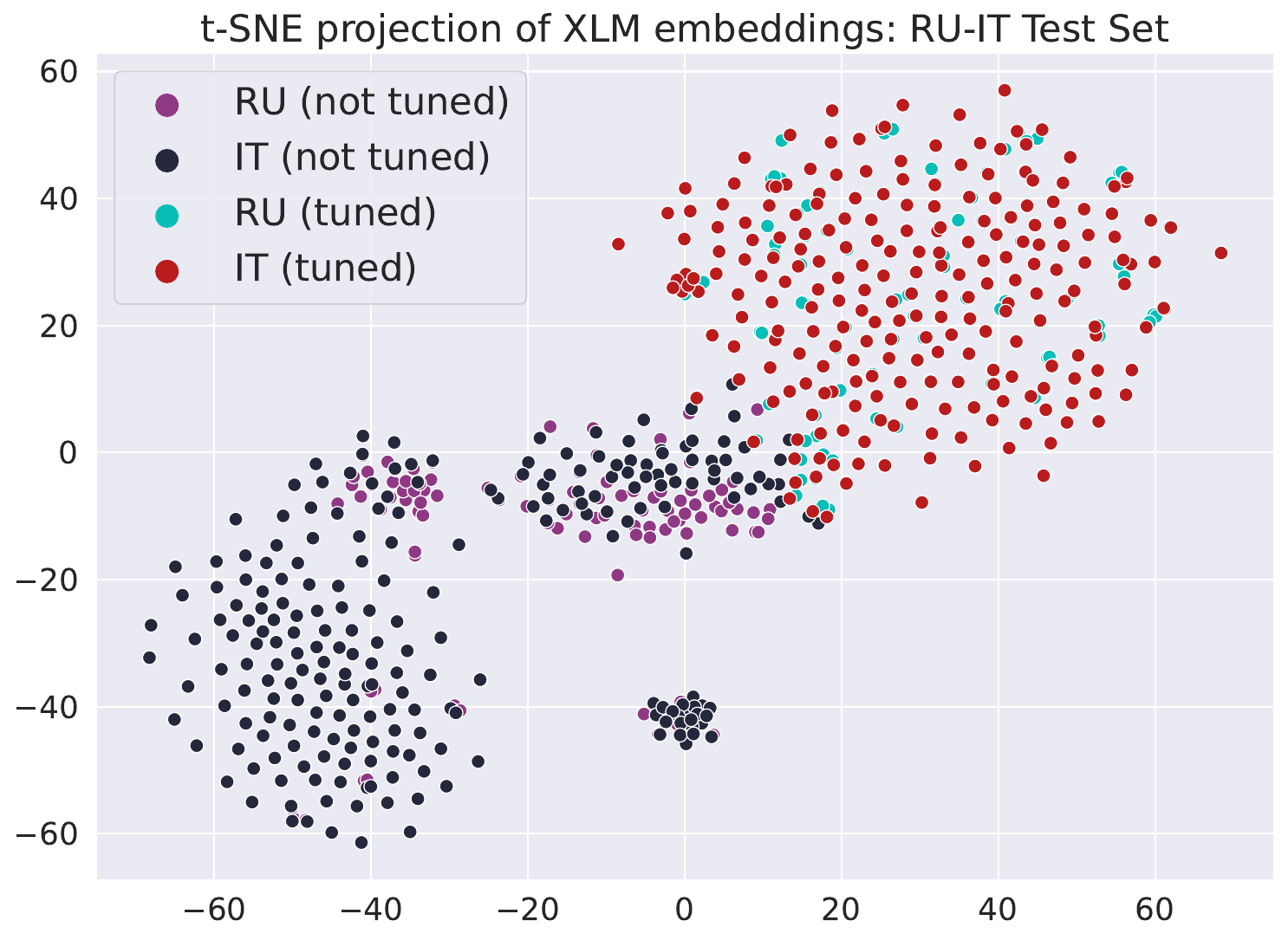}
    \caption{A t-SNE visualization of word embeddings generated from the mPLM for words in the RU-IT pair test set, before and after the prompt-based finetuning.}
\end{minipage}%
\end{figure*}

\end{document}